\documentclass[acmtog,authorversion]{acmart} % for final
\acmSubmissionID{208}

\usepackage{booktabs} % For formal tables
\usepackage{multirow}
\usepackage{tabularx}
\usepackage{wrapfig}
\usepackage{mathtools}

\usepackage[ruled]{algorithm2e} % For algorithms

\SetAlFnt{\small}
\SetAlCapFnt{\small}
\SetAlCapNameFnt{\small}
\SetAlCapHSkip{0pt}

\setcopyright{acmlicensed}
\acmJournal{TOG}
\acmYear{2021}\acmVolume{40}\acmNumber{4}\acmArticle{151}\acmMonth{8}
\acmDOI{10.1145/3450626.3459766}

\begin{document}
\def\negativevspace{}	
\def\ie{\emph{i.e.}}
\def\eg{\emph{e.g.}}
\def\etal{{\em et al.}}
\def\etc{{\em etc.}}
\newcolumntype{C}[1]{>{\centering\arraybackslash}p{#1}}
	
% Title portion
\title{SP-GAN: Sphere-Guided 3D Shape Generation and Manipulation}

\author{Ruihui Li}
\author{Xianzhi Li}
\author{Ka-Hei Hui}
\author{Chi-Wing Fu}
\affiliation{%
	\institution{The Chinese University of Hong Kong}
    \city{Hong Kong}
    \country{China}
}
\renewcommand\shortauthors{Li et al.}

\begin{teaserfigure}
\centering
\includegraphics[width=0.925\linewidth]{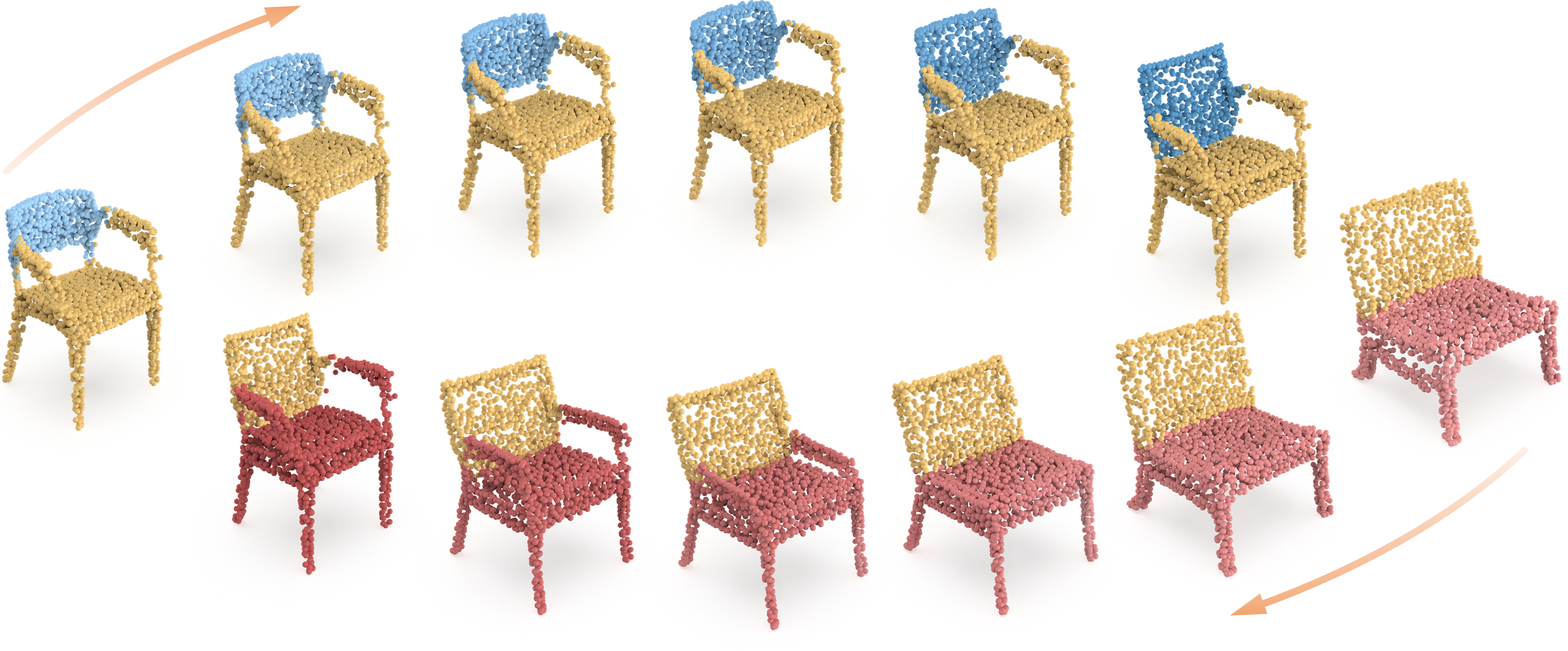}
\vspace{-4.5mm}
\caption{SP-GAN not only enables the generation of diverse and realistic shapes as point clouds with fine details
(see the two chairs on the left and right) but also embeds a dense correspondence across the generated shapes, thus facilitating part-wise interpolation between user-selected local parts in the generated shapes.
Note how the left chair's back (blue part in top arc) and the right chair's legs (red part in bottom arc) morph in the two sequences.
}
\label{fig:teaser}
\end{teaserfigure}

\begin{abstract}
We present SP-GAN, a new unsupervised sphere-guided generative model for direct synthesis of 3D shapes in the form of point clouds.
Compared with existing models, SP-GAN is able to synthesize diverse and high-quality shapes with fine details and promote controllability for part-aware shape generation and manipulation, yet trainable without any parts annotations.
In SP-GAN, we incorporate a global prior (uniform points on a sphere) to spatially guide the generative process and attach a local prior (a random latent code) to each sphere point to provide local details.
The key insight in our design is to disentangle the complex 3D shape generation task into a global shape modeling and a local structure adjustment, to ease the learning process and enhance the shape generation quality.
Also, our model forms an implicit dense correspondence between the sphere points and points in every generated shape,
enabling various forms of structure-aware shape manipulations such as part editing, part-wise shape interpolation, and multi-shape part composition, etc., beyond the existing generative models.
Experimental results, which include both visual and quantitative evaluations, demonstrate that our model is able to synthesize diverse point clouds with fine details and less noise, as compared with the state-of-the-art models.
\end{abstract}




%\keywords{3D point clouds, unsupervised generative model, dense correspondence, structure-aware generation and manipulation}

\maketitle

\section{Introduction}
\label{sec:intro}

A challenging problem in 3D shape creation is how to build generative models to synthesize new, diverse, and realistic-looking 3D shapes, while having structure-aware generation and manipulation.
One solution is to decompose shapes into parts~\cite{mo2019structurenet,wu2020pq} and learn to compose new shapes according to part-level relations.
With parts correspondence, these approaches can further enable various forms of structure-aware manipulation; however, the granularity of the shape generation is part-based and it depends on the availability and quality of the parts annotations.

Another feasible solution is to design deep generative models to characterize the shape distribution and learn to directly generate shapes in an unsupervised manner.
Prior researches generate shapes represented as 3D point clouds, typically by learning to map random latent code to point clouds, via auto-regressive models~\cite{sun2020pointgrow}, flow-based models~\cite{yang2019pointflow,klokov2020discrete,kim2020softflow}, and generative adversarial nets (GAN)~\cite{achlioptas2018learning,shu20193d,hui2020progressive}.
Though substantial progress has been made, the problem is still very challenging, due to the diverse shape variations and the high complexity in 3D space.
Hence, existing generative models often struggle with the fine details and tend to produce noisy point samples in the generated shapes.
Also, existing models lack structure controllability for part-aware generation and manipulation; it is because the learned mapping from a single latent code merely characterizes the overall shape variation, so it is hard to obtain a
plausible correspondence between
the parts in generated shapes and the dimensions in latent code, as well as across different shapes generated by the learned mapping.

This paper presents a new GAN model called SP-GAN, in which we use a Sphere as Prior to guide the direct generation of 3D shapes (in the form of point clouds).
Our new approach goes beyond generating diverse and realistic shapes, in which we are able to {\em generate point clouds with finer details and less noise\/}, as compared with the state-of-the-arts.
Importantly, our design facilitates {\em controllability in the generative process\/}, since our model implicitly embeds a dense correspondence between the generated shapes, and training our model is {\em unsupervised\/}, without requiring any parts annotations.
By this means, we can perform {\em part-aware generation and manipulation of shapes\/}, as demonstrated by the results shown in Figure~\ref{fig:teaser}.

\begin{figure}[t]
\centering
\includegraphics[width=0.99\linewidth]{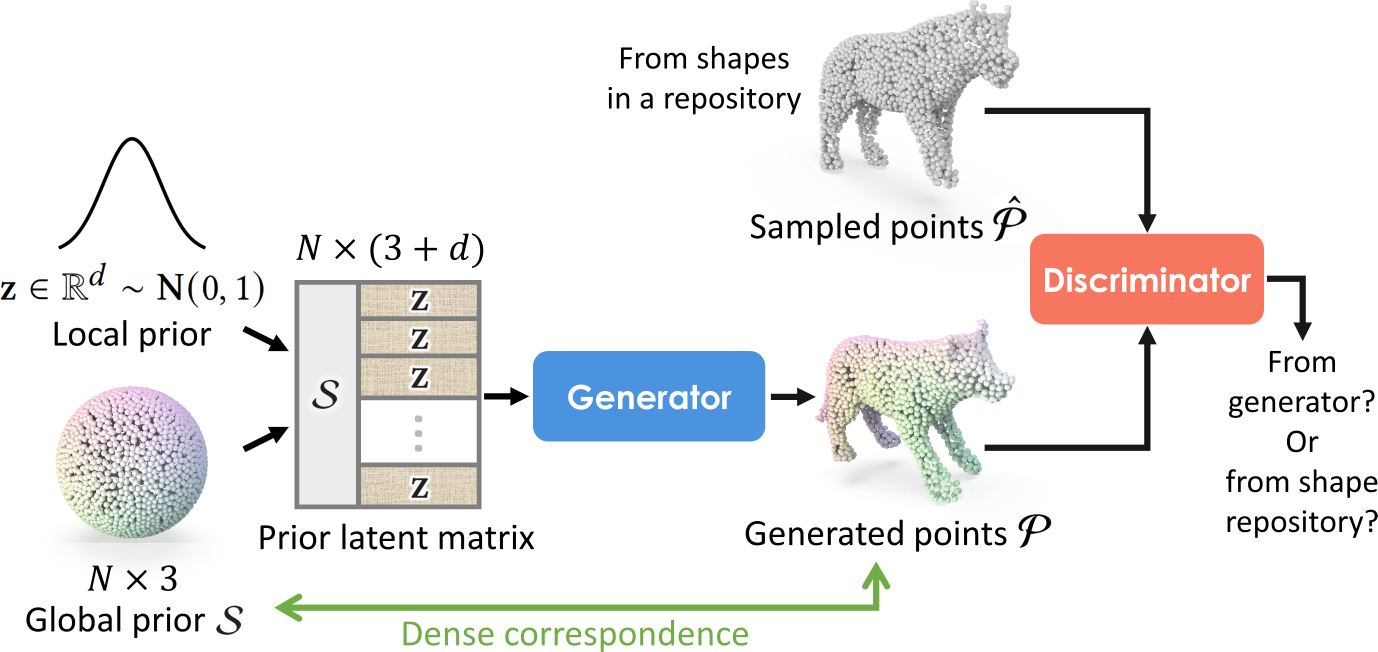}
\vspace{-1mm}
\caption{An overview of SP-GAN.
Its input is a {\em prior latent matrix\/} with (i) a random latent code $\mathbf{z}$, and (ii) a unit 3D sphere $\mathcal{S}$, which is represented as $N$ points evenly-distributed on the sphere.
In SP-GAN, $\mathcal{S}$ provides an {\em unbiased spatial guidance\/} to the generator for it to learn to synthesize point cloud $\mathcal{P}$ from $\mathbf{z}$ with finer details and less noise.
Also, SP-GAN implicitly embeds a {\em dense correspondence\/} between $\mathcal{S}$ and $\mathcal{P}$ (see their colors above), thus promoting structure-aware shape generation and manipulation.}
\label{fig:overview}
\vspace{-1.5mm}
\end{figure}

Figure~\ref{fig:overview} illustrates the key design in SP-GAN.
Instead of having a single latent code as input like the conventional generative models, we design our generator with two {\em decoupled\/} inputs:
(i) {\em a global prior\/} $\mathcal{S}$, which is a fixed 3D point cloud in the form of a unit sphere, to provide an isotropic (unbiased) spatial guidance to the generative process; and
(ii) {\em a local prior\/} $\mathbf{z}$, which is a random latent code to provide local details.
We pack these two inputs into a {\em prior latent matrix\/} by attaching a latent code $\mathbf{z}$ to every point in $\mathcal{S}$, as illustrated in Figure~\ref{fig:overview}.
By this means, the generative process starts from a shared global initialization (\eg, a common ground), yet accommodating the spatial variations, such that every point moves towards its desired location for forming the shape.
A key insight behind our design is that we {\em formulate the generation task as a transformation\/} and {\em disentangle the complex 3D shape generation task\/} into
(i) a global shape modeling (\eg, chair) and
(ii) a local structure adjustment (\eg, chair's legs).
Such a decoupling scheme eases the learning process and enhances the quality of the generated shapes.

Another important consequence is that our new model facilitates structure controllability in the generative process, {\em through an implicit dense correspondence between the input sphere and the generated shapes\/}.
Metaphorically, the sphere provides a common working space (like the canvas in painting) for shape generation and manipulation; painting on a certain area on the sphere naturally manipulates the corresponding part in different generated shapes.
So, if we modify the latent vectors associated with specific points on $\mathcal{S}$ while keeping others unchanged, we can manipulate local structures for the associated parts in the shape. Also, since parts are connected geometrically and semantically with one another, changing one part may cause slight changes in some other parts for structural compatibility; see Figure~\ref{fig:teaser} (bottom).
Furthermore, the dense correspondence extends {\em across all the generated shapes\/} produced from the generator, with $\mathcal{S}$ serving as a proxy.
Hence, we can interpolate the latent code of different generated shapes to morph between shapes in a shape-wise or part-wise fashion.
We shall show various structure-aware shape manipulations in Section~\ref{sec:generation}.
Further, the experimental evaluations confirm that SP-GAN is able to generate diverse and high-quality point clouds, compared with the state-of-the-art generative models.

\section{Related Work}
\label{sec:rw}

3D shape generation based on deep neural networks has attracted immense research interest.
Broadly speaking, it relates to many areas in graphics and vision, for example, 3D shape reconstruction from various forms of input (2D RGB images~\cite{wu2017marrnet,knyaz2018image,zheng2019deephuman}, 2.5D depth images~\cite{wu20153d,yang2018dense,waechter2017virtual,li2017modeling,guo2017real}, and 3D scans~\cite{hanocka2020point2mesh,yuan2018pcn,aberman2017dip}), shape transform~\cite{yin2019logan,yin2018p2p},etc.
In this work, we focus on 3D shape generation, specifically in the context of generating new and diverse shapes that are not necessarily in the given shape repository but yet look realistic when compared with the existing shapes in the repository.

In general, 3D shapes can be generated in a \emph{part-conditioned} or \emph{unconditioned} manner.
Part-conditioned means that we employ an additional dataset with parts labels to train a part-wise variational autoencoder to encode each predefined (or learned) part into a latent distribution~\cite{mo2019structurenet,mo2020pt2pc,wu2020pq,dubrovina2019composite}, so that new shapes can be generated by sampling from the distributions of parts and composing them.
On the other hand, unconditioned means that we directly synthesize 3D shapes from a random distribution, so the generation process has full freedom and the generated samples are not limited by the part-annotation data or any pre-defined structure relation.

Our SP-GAN model is unconditional, so we further review works on unconditional approaches from now on. With advances in direct 3D point cloud processing using deep neural networks, as inspired by pioneering works such as~\cite{qi2016pointnet,qi2017pointnet++,wang2019dynamic},~\etc, several new approaches were proposed recently to generate 3D shapes in the form of point clouds.
These methods can be roughly classified into autoregressive-based, flow-based, and GAN-based.

%%%%%%%%%%%%%%%%%%%%%%%%%%%%%%%%%%
%\noindent\textbf{Autoregressive-based generative approach}
%\vspace{.05in}\noindent\textbf{Autoregressive-based generative approach}
%\paragraph{}
%\paragraph{Autoregressive-based generative approach}
%\paragraph{
\vspace{.05in}
{\em Autoregressive-based generative approach\/}
models the joint 3D spatial distribution of points in a point cloud.
Specifically, Sun~\etal~\shortcite{sun2020pointgrow} propose PointGrow to estimate the point coordinate distributions from the training shapes.
During the generative phase, points are sampled one-by-one based on the estimated probability distributions given the previously-generated points.
However, due to the iterative property intrinsic to autoregressive models, the model cannot scale well with the size of the point cloud.

%%%%%%%%%%%%%%%%%%%%%%%%%%%%%%%%%%

%\vspace{.05in}\noindent\textbf{Flow-based generative approach} 
\vspace{.05in}
{\em Flow-based generative approach\/} learns to model the distribution of points in a shape mainly by an invertible parameterized transformation of the points.
In the generative phase, the approach samples points from a given generic prior (\eg, a Gaussian distribution) and moves them to the target shape using the learned parameterized transformation.
For example, Yang~\etal~\shortcite{yang2019pointflow} generate point clouds from a standard 3D Gaussian prior based on continuous normalizing flows.
Klokov~\etal~\shortcite{klokov2020discrete} relieve the computation by formulating a model using discrete normalizing flows with affine coupling layers~\cite{dinh2016density}.
To better characterize the data distribution, Kim~\etal~\shortcite{kim2020softflow} propose to learn a conditional distribution of the perturbed training shapes.
Cai~\etal~\shortcite{cai2020learning} model the gradient of the log-density field of shapes and generate point clouds by moving the sampled points towards the high-likelihood regions.

While substantial progress has been made, the invertibility constraint in flow-based approach unavoidably limits the representation capability of the models.
Also, the learned parameterized transformation can be regarded as a rough estimation of the averaged training data distribution, so the generated point samples tend to be blurry and noisy, as we shall show later in Section~\ref{sec:evaluation}.
%%%%%%%%%%%%%%%%%%%%%%%%%%%%%%%%%%

%\vspace{.05in}\noindent\textbf{GAN-based generative approach} 
\vspace{.05in}
{\em GAN-based generative approach\/} explores adversarial learning to train the shape generation model with the help of a discriminator.
Achlioptas~\etal~\shortcite{achlioptas2018learning} introduce the first set of deep generative models to produce point clouds from a Gaussian noise vector, including an r-GAN that operates on a raw point cloud input and an l-GAN that operates on the bottleneck latent variables of a pre-trained autoencoder.
To overcome the redundancy and structural irregularity of point samples, Ramasinghe~\etal~\shortcite{ramasinghe2019spectral} propose Spectral-GANs to synthesize shapes using a spherical-harmonics-based representation.
Shu~\etal~\shortcite{shu20193d} propose tree-GAN to perform graph convolutions in a tree and Gal~\etal~\shortcite{gal2020mrgan} recently extend it into a multi-rooted version.
Hui~\etal~\shortcite{hui2020progressive} design a progressive deconvolution network to generate 3D point clouds, while Arshad~\etal~\shortcite{arshad2020progressive} create a conditional generative adversarial network to produce dense colored point clouds in a progressive manner.

Compared with the above GAN-based approaches, which attempt to synthesize point clouds from only a single latent code, we incorporate a fixed prior shape to guide the generative process in the form of a disentanglement model. Our model not only produces high-quality outputs with fine details but also promotes controllability for structure-aware shape generation and manipulation.

%\vspace{.05in}\noindent\textbf{Other 3D representations}, 

\vspace{.05in}
Other 3D representations such as voxel grid~\cite{wu2016learning,smith2017improved}, implicit function~\cite{chen2019learning,deng2020deformed}, and deformable mesh~\cite{sinha2017surfnet,groueix2018papier,wang2018pixel2mesh} are also explored for shape generation in recent years.
Voxels are natural extensions of image pixels, allowing state-of-the-art techniques to migrate from image-space processing to 3D shape processing.
However, 3D voxel grids suffer from large memory consumption and low-resolution representation, so the generated shapes contain only coarse structures but not fine details.
Implicit-function-based methods~\cite{mescheder2019occupancy,park2019deepsdf} are flexible but computationally expensive, since a large amount of points (e.g., $128^3$) must be sampled to get a surface from the representation, leading to a huge computational resource demand.
Mesh-based methods, such as Pixel2Mesh~\cite{wang2018pixel2mesh}, learn to deform a surface template to form the target shape given an input image. It is, however, hard to generate and interpolate shapes with arbitrary (and different) genus.  Also, the reconstruction is conditional, meaning that it requires reference images and ground-truth 3D shapes, instead of being unconditional as in SP-GAN.

Using points for shape generation has several advantages.
First, points are a generic representation for 3D shapes, without constraining the topologies and genus. Also, points are easy to use for shape manipulation and interpolation.
Further, points are naturally acquired by scanning without tedious post-processing.

%\vspace{.05in}\noindent\textbf{Other related works.} 
%\paragraph{Other related works}
\vspace{.05in}
Contemporarily, Deng~\etal~\shortcite{deng2020deformed} propose a deformed implicit field representation for modeling dense correspondence among shapes.
It learns an implicit-field-based template for each category and deforms the template towards each given shape to build the shape correspondence explicitly. In SP-GAN, we adopt a general ``sphere prior'' to enable high-quality shape synthesis, in which the dense correspondence is embedded implicitly and automatically via our novel design.
Similarly, we noticed that P2P-Net~\cite{yin2018p2p} attaches an independent noise vector to each point feature for providing more freedom to the displacements of individual points.
Differently, we attach noise latent code to each sphere point for providing local details and the point relations together define the global structure of the generated shapes.

%%%end of section 
\section{Overview}
\label{sec:overview}

\begin{figure}[!t]
\centering
\includegraphics[width=0.92\linewidth]{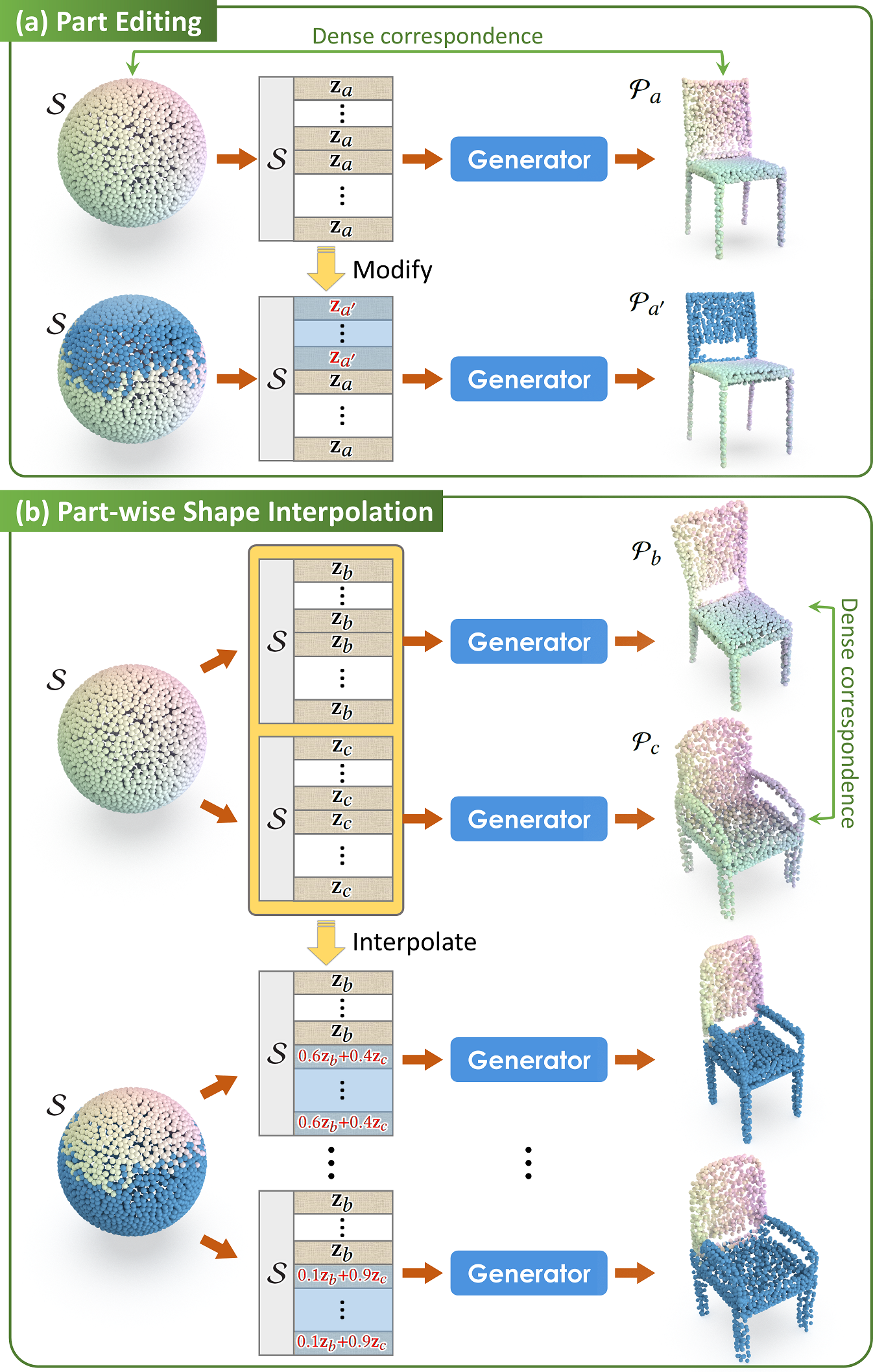}
\vspace*{-2.5mm}
\caption{SP-GAN establishes a dense correspondence implicitly between sphere $\mathcal{S}$ and all the generated shapes, with sphere $\mathcal{S}$ as a proxy.
By this model, we can modify or interpolate specific latent code in the prior latent matrix, e.g., for part editing (a) and part-wise shape interpolation (b). }
\label{fig:manipulate_pipeline}
\vspace*{-2.5mm}
\end{figure}

\begin{figure*}[t]
\centering
\includegraphics[width=0.99\linewidth]{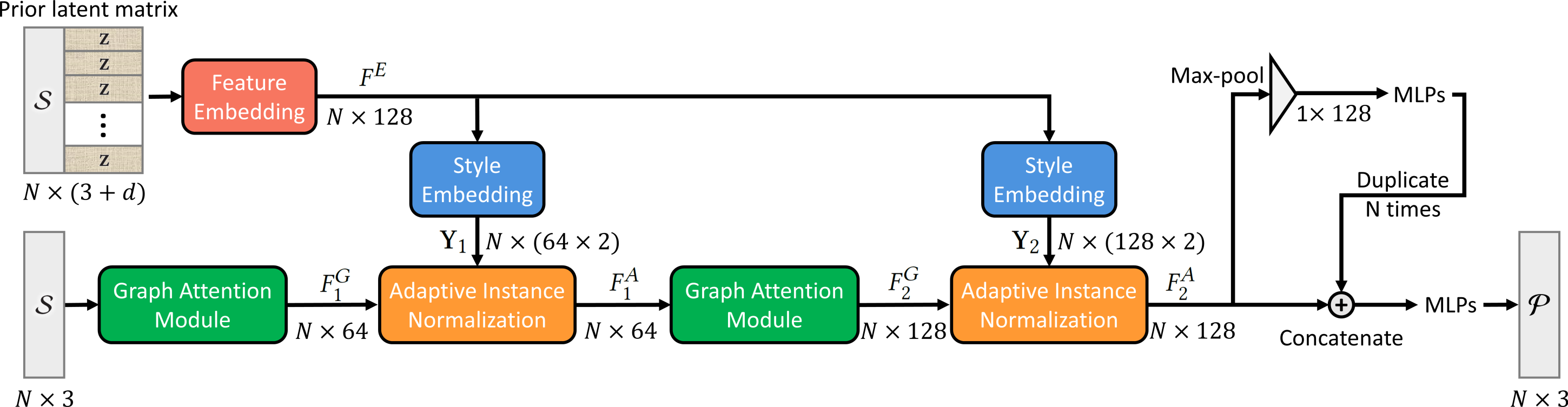}
\vspace*{-1.5mm}
\caption{Architecture of the generator in SP-GAN.
With the prior latent matrix as input, the generator first feeds sphere points $\mathcal{S}$ to a graph attention module (green box; see Figure~\ref{fig:gcn}) to extract the point-wise feature map $F^G_1$.
On the top branch, we use a nonlinear feature embedding (red box) to extract local style features $F^E$ from the prior latent matrix.
Then, to fuse the local styles in $F^E$ with the spatial features in $F^G_1$, we propose to embed local styles $\mathbf{Y}_1$ from $F^E$ (blue box) and then use adaptive instance normalization (orange box) to produce $F^A_1$ with richer local details.
We repeat the process with another round of style embedding and normalization, and then follow PointNet~\cite{qi2016pointnet} to reconstruct point cloud $\mathcal{P}$ from the embedded feature map $F^A_2$.}
\label{fig:generator}
\vspace*{-1.5mm}
\end{figure*}

%%%%%%%%%%%%%%%%%%%%%%%%%%%%%%%%%%%%%%%%%%

The basic model of SP-GAN is illustrated in Figure~\ref{fig:overview}, with shapes represented as point clouds like the previous works we discussed.
In SP-GAN, the generator consumes two {\em decoupled inputs\/}:
a global prior $\mathcal{S}$, which contains the 3D coordinates of $N$ points uniformly on a unit sphere, and a local prior $\mathbf{z}$, which is a $d$-dimensional random latent vector, with each element randomly sampled from a standard normal distribution.
Very importantly, we pack one latent vector $\mathbf{z}$ with each point in $\mathcal{S}$ to form the {\em prior latent matrix\/} as the generator input; see again Figure~\ref{fig:overview}.
During the training, we use the generator to synthesize point cloud $\mathcal{P} \in \mathbb{R}^{N \times3}$ from $\mathcal{S}$ and $\mathbf{z}$; besides, we sample another point cloud $\mathcal{\hat{P}} \in \mathbb{R}^{N \times3}$ from shape in a given 3D repository.
So, when we train SP-GAN, the discriminator should learn to differentiate $\mathcal{P}$ and $\mathcal{\hat{P}}$, while the generator should learn to produce $\mathcal{P}$ that looks like $\{\mathcal{\hat{P}}\}$ from the 3D repository.

During the testing, we randomize a latent code and pack it with sphere $\mathcal{S}$ into a prior latent matrix, and feed it to the trained generator to produce a new shape; see $\mathcal{P}_a$ in Figure~\ref{fig:manipulate_pipeline}(a).
Thanks to the sphere proxy, training SP-GAN creates an {\em implicit association\/} between points in $\mathcal{S}$ and points in the generated shape.
This is a {\em dense point-wise correspondence\/}, as illustrated by the smoothly-varying point colors on the sphere and on the generated shape shown in Figure~\ref{fig:manipulate_pipeline}(a).
Also, this dense correspondence extends across all the shapes produced from the generator;
see the colors of the points in generated shapes $\mathcal{P}_b$ and $\mathcal{P}_c$ in Figure~\ref{fig:manipulate_pipeline}(b).
With this important property, SP-GAN facilitates various forms of structure- and part-aware shape generation and manipulation.
Below, we first describe two basic cases and more can be found later in Section~\ref{sec:generation}:
\begin{itemize}
\item
\emph{Part editing}.
The latent code associated with each point in sphere $\mathcal{S}$ is a local prior.
If we change the latent code for some of the points, we can make local changes on the associated part in the generated shape.
Figure~\ref{fig:manipulate_pipeline}(a) shows an example, in which we modify the latent code of the points associated with the chair's back (in blue) from $\mathbf{z}_{a}$ to $\mathbf{z}_{a'}$.
By then, the generator produces a new chair $\mathcal{P}_{a'}$, whose back follows $\mathbf{z}_{a'}$ and other parts are slightly adjusted for compatibility with the newly-modified back.
\item
\emph{Part-wise interpolation}.
Further, the dense correspondence enables us to interpolate between the latent code of different generated shapes to morph shapes in part-wise fashion.
This manipulation cannot be achieved by any existing unconditional generative models for 3D point clouds.
Figure~\ref{fig:manipulate_pipeline}(b) shows an example, in which we interpolate the latent code associated with the chair's lower part (marked as blue) between $\mathbf{z}_b$ and $\mathbf{z}_c$ by setting different interpolation weights.
With this change, the generator produces a set of new chairs, in which their lower parts morph from $\mathcal{P}_b$ to $\mathcal{P}_c$.
\end{itemize}

%%%end of section

\section{SP-GAN}
\label{sec:architecture}

In this section, we first present the architecture design of the generator and discriminator networks in SP-GAN, and then present the training and implementation details.

%%%%%%%%%%%%%%%%%%%%%%%%%%%%%%%%%%%%%%%%%%%%%%%%%%

\subsection{Generator}

Figure~\ref{fig:generator} shows the architecture of the generator in SP-GAN, which produces point cloud $\mathcal{P} \in \mathbb{R}^{N \times 3}$ from sphere $\mathcal{S}$ and a prior latent matrix.
Inside the generator, $\mathbf{z}$ introduces diverse local styles and fine details into point features, whereas $\mathcal{S}$ serves as a prior shape for feature extraction and guides the generative process.
Particularly, $\mathcal{S}$ also allows us to employ graph-related convolutions with spatial correlation for feature extraction.
This is very different from existing works on 3D point cloud generation, which take only a single latent code as input; hence, existing models can only use fully-connected layers at early stages and require relatively large amount of learnable parameters, yet having limited expressiveness.
\begin{figure}[!t]
	\centering
	\includegraphics[width=0.95\linewidth]{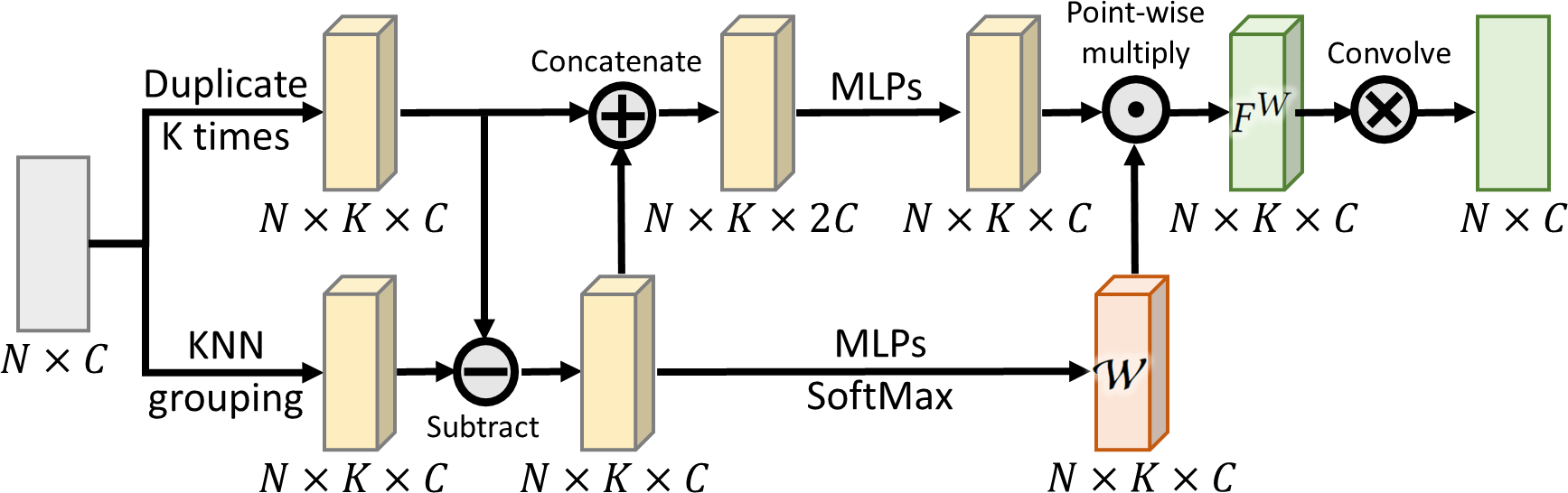}
	\vspace*{-1.5mm}
	\caption{Architecture of the graph attention module. Besides adopting the basic module (light yellow) in DGCNN~\cite{wang2019dynamic}, instead of equally treating the neighbor features, we regress weights $\mathcal{W}$ to obtain the weighted feature map $F^W$ by considering the relations among the $K$ point neighbors.}
	\label{fig:gcn}
	\vspace*{-1.5mm}
\end{figure}

To elaborate on how our generator works, let's start with the bottom branch shown in Figure~\ref{fig:generator}.
First, the generator employs a graph attention module (to be detailed soon) to extract a point-wise feature map $F^G_1 \in \mathbb{R}^{N \times 64}$ from $\mathcal{S}$.
Now, to generate 3D shapes, an important question is how to insert the structure variations and local styles brought from latent code $\mathbf{z}$ into feature map $F^G_1$.
Inspired by style transfer~\cite{dumoulin2016learned,karras2019style}, we propose to use an adaptive instance normalization layer (the left orange box in Figure~\ref{fig:generator}) by modulating the mean and variance of the feature vectors in $F^G_1$.
Specifically, as shown in the top branch, we employ a non-linear feature embedding, which is implemented using multi-layer perceptrons (MLPs), to transform the prior latent matrix into a high-level feature map $F^E \in \mathbb{R}^{N \times 128}$.
Then, we use a style embedding (the left blue box in Figure~\ref{fig:generator}), which is also implemented using MLPs, to specialize $F^E$ into styles $\mathbf{Y}_1=(\mathbf{Y}_1^s, \mathbf{Y}_1^b)$, where $\mathbf{Y}_1^s \in \mathbb{R}^{N \times 64}$ and $\mathbf{Y}_1^b \in \mathbb{R}^{N \times 64}$ control the scale and bias, respectively, when normalizing each point feature vector in $F^G_1$.
Thus, the adaptive instance normalization operation becomes
\begin{equation}
\label{equ:adaIN}
F_1^A(i) = \mathbf{Y}_1^s(i) \cdot \frac{F^G_1(i)-\mu(F^G_1(i))}{\sigma(F^G_1(i))}+\mathbf{Y}_1^b(i), i \in [1, ..., N],
\end{equation}
where
$F^G_1(i)$ is the $i$-th point feature vector in $F^G_1$,
$F_1^A(i)$ is the corresponding feature vector after the normalization, and
$\mu(\cdot)$ and $\sigma(\cdot)$ compute the mean and standard deviation across the spatial axes of its argument, which is a feature vector.
In general, Eq.~\eqref{equ:adaIN} allows us to encode the learned per-point style,~\ie, $(\mathbf{Y}_1^s(i), \mathbf{Y}_1^b(i))$, into the associated embedded feature vector $F^G_1(i)$, so that the final adjusted $F^A_1$ contains richer local details.

To further enrich the feature embedding, we pass $F^A_1$ to another set of graph attention module and adaptive instance normalization to obtain $F^G_2$ and $F^A_2$, respectively.
As shown on the right side of Figure~\ref{fig:generator}, we further regress the output 3D point cloud $\mathcal{P}$ by following PointNet~\cite{qi2016pointnet} to reconstruct the 3D coordinates,~\ie, by concatenating $F^A_2$ with the duplicated global feature vectors.
Please refer to~\cite{qi2016pointnet} for the details.

%\vspace{.05in}\noindent\textbf{Graph attention module.} \
\paragraph{Graph attention module}
Next, we elaborate on the design of the graph attention module; see Figure~\ref{fig:gcn} for an illustration of its architecture. Here, we adopt the basic module (marked in light yellow) from DGCNN~\cite{wang2019dynamic} and make our adjustments to further account for the relationships among the $K$ neighbors in the feature space.
Instead of taking the neighboring features equally, we regress weights $\mathcal{W} \in \mathbb{R}^{N \times K \times C}$ (orange box) and employ point-wise multiplication to obtain the weighted neighbouring feature map $F^W$.
Lastly, the output $N \times C$ feature map is obtained by applying a convolution of a kernel size of $1\times K$ to $F^W$.

%%%%%%%%%%%%%%%%%%%%%%%%%%%%%%%%%%%%%%%%%%%%%%%%%%%

\subsection{Discriminator}
\begin{figure}[!t]
	\centering
	\includegraphics[width=0.9\linewidth]{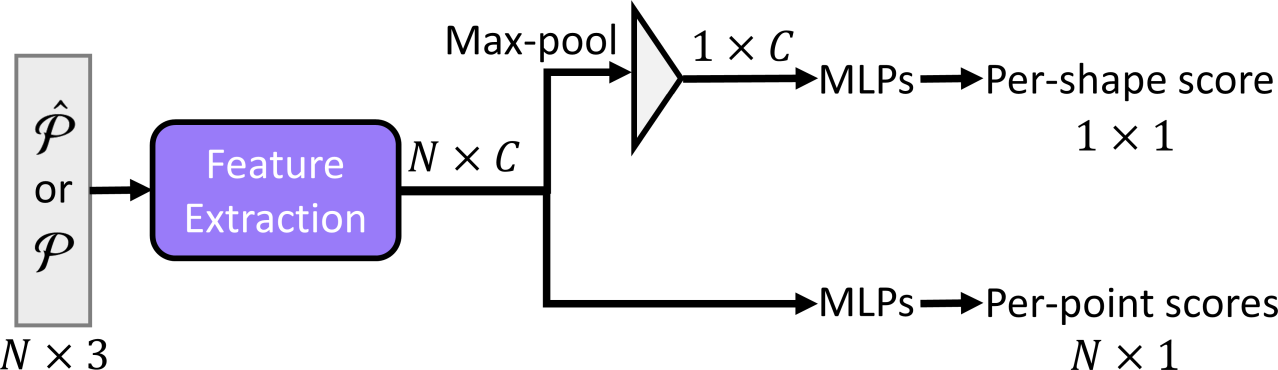}
	\vspace*{-1.5mm}
	\caption{Architecture of the discriminator in SP-GAN.
	To classify the input point cloud as being produced by the generator or sampled from a given 3D repository, we predict not only a per-shape score (top branch) but also per-point scores (bottom branch) to encourage fine-grained analysis.}
	\label{fig:discriminator}
	\vspace*{-1.5mm}
\end{figure}

Figure~\ref{fig:discriminator} shows the architecture of the discriminator in SP-GAN.
Its input is either a point cloud $\mathcal{P}$ produced by the generator or a point cloud $\mathcal{\hat{P}}$ sampled from shape in a given 3D repository.
From the $N$ points in $\mathcal{P}$ or $\mathcal{\hat{P}}$, a conventional discriminator would learn a $1 \times C$ global feature vector for the whole shape and predict a single score that indicates the source of the point cloud.
Considering that the ``realism'' of an input can be explored by looking into its local details, as inspired by~\cite{schonfeld2020u}, we extend the conventional discriminator to additionally perform classifications on a per-point basis,~\ie, by predicting a score for each point in the input.

As shown in Figure~\ref{fig:discriminator}, after we extract point features from the input, we use the top branch to predict a per-shape score and the bottom branch to predict per-point scores.
In this way, the discriminator can effectively regularize both global and local variations in the input point cloud.
Correspondingly, we can then encourage the generator to focus on both global structures and local details, such that it can better synthesize point clouds to fool this more powerful discriminator.
Note that in our implementation, we adopt PointNet~\cite{qi2016pointnet} as the backbone for the feature extraction, which is the violet box shown in Figure~\ref{fig:discriminator}.

%%%%%%%%%%%%%%%%%%%%%%%%%%%%%%%%%%%%%%%%%%%%%%%%%%
\subsection{Training and Implementation details}
%Training and Losses, and all Implementation details
\label{subsec:train}

%\vspace{.05in}\noindent\textbf{Loss functions.} \
\paragraph{Loss functions}
SP-GAN can be trained end-to-end using one loss for generator G and another loss for discriminator D.
In this work, we design the two losses based on the least squares loss~\cite{mao2017least}, which is originally formulated by minimizing the following competing objectives in an alternating manner:
\begin{eqnarray}
\label{equ:ori_gen}
\mathcal{L}_G&=&\frac{1}{2}[D(\mathcal{P})-1]^2 \\
\text{and} \ \
\label{equ:ori_dis}
\mathcal{L}_D&=&\frac{1}{2}[(D(\mathcal{P})-0)^2 + (D(\mathcal{\hat{P}})-1)^2],
\end{eqnarray}
where $D(\mathcal{P})$ and $D(\mathcal{\hat{P}})$ are the confidence values predicted by the discriminator on $\mathcal{P}$ and $\mathcal{\hat{P}}$, respectively, and $\mathcal{L}_G$ and $\mathcal{L}_D$ are the loss for training the generator and discriminator, respectively.

Since our discriminator outputs two types of predictions,~\ie, per-shape score and per-point scores.
Hence, we model the discriminator loss $\mathcal{L}_D$ as a summation of shape loss $\mathcal{L}_D^{\text{shape}}$ and point loss $\mathcal{L}_D^{\text{point}}$:
\begin{eqnarray}
\mathcal{L}_D & = & \mathcal{L}_D^{\text{shape}} + \lambda \mathcal{L}_D^{\text{point}},
\label{equ:discriminator}
\\
\mathcal{L}_D^{\text{shape}} & = & \frac{1}{2}[(D(\mathcal{P})-0)^2 + (D(\mathcal{\hat{P}})-1)^2],
\label{equ:shape}
\\
\mathcal{L}_D^{\text{point}} & = & \frac{1}{2N}\sum_{i=1}^{N}[(D(p_i)-0)^2 + (D(\hat{p}_i)-1)^2].
\label{equ:point}
\end{eqnarray}
where $\lambda$ is a balance parameter;
$p_i$ and $\hat{p}_i$ are the $i$-th point in $\mathcal{P}$ and $\mathcal{\hat{P}}$, respectively; and
$\mathcal{L}_D^{\text{point}}$ is computed by averaging the predictions of all the points.

Correspondingly, the objective for training the generator becomes
\begin{equation}
\label{equ:generator}
\mathcal{L}_G = \frac{1}{2}[D(\mathcal{P})-1]^2 +\beta \frac{1}{2N}\sum_{i=1}^{N}[D(p_i)-1]^2,
\end{equation}
where $\beta$ is a weight to balance the terms.

%%%%%%%%%%%%%%%%%%%%%%%%%%%%%%%%%%%%%%%

%\vspace{.05in}\noindent\textbf{Network training.} \
\paragraph{Network training}
Overall, we train SP-GAN by alternatively optimizing the discriminator using Eq.~\eqref{equ:discriminator} and the generator using Eq.~\eqref{equ:generator}.
Note also that, we keep sphere $\mathcal{S}$ unchanged during the training and randomly sample a latent code $\mathbf{z}$ from a standard normal distribution in each of the training iterations.

%%%%%%%%%%%%%%%%%%%%%%%%%%%%%%%%%%%%%%%
%\vspace{.05in}\noindent\textbf{Network inference.} \
\paragraph{Network inference}
During the generation phase, we only need to use the trained generator to produce a point cloud, while keeping the same $\mathcal{S}$ as in the training.
With different randomly-sampled $\mathbf{z}$ as inputs, the generator can produce diverse point cloud outputs.

%%%%%%%%%%%%%%%%%%%%%%%%%%%%%%%%%%%%%%%

%\vspace{.05in}\noindent\textbf{Implementation details.} \
\paragraph{Implementation details}
We implement our framework using PyTorch and train it on a single NVidia Titan Xp GPU using the Adam optimizer.
We use a learning rate of $10^{-4}$ to train both the generator and discriminator networks,
and follow the alternative training strategy in~\cite{goodfellow2014generative} to train each of them for 300 epochs.
In both networks, we use LeakyReLU as a nonlinearity activation function.
In the last layer of the generator, we use $\tanh$ as the activation function.
Also, we set $K$$=$$20$ to extract point neighborhood.
Our code is available for download on GitHub\footnote{https://github.com/liruihui/SP-GAN}.
%%%end of section

\section{Shape Generation and Manipulation}
\label{sec:generation}

This section presents various forms of shape generation, manipulation, and analysis that are enabled by SP-GAN.

%%%%%%%%%%%%%%%%%%%%%%%%%%%%%%%%%%%%%%%%%%%%%%%%%%%%%%%%%%%%%%%

\subsection{Shape Generation}
\label{subsec:vis}
\begin{figure*}
	\centering
	\includegraphics[width=0.95\linewidth]{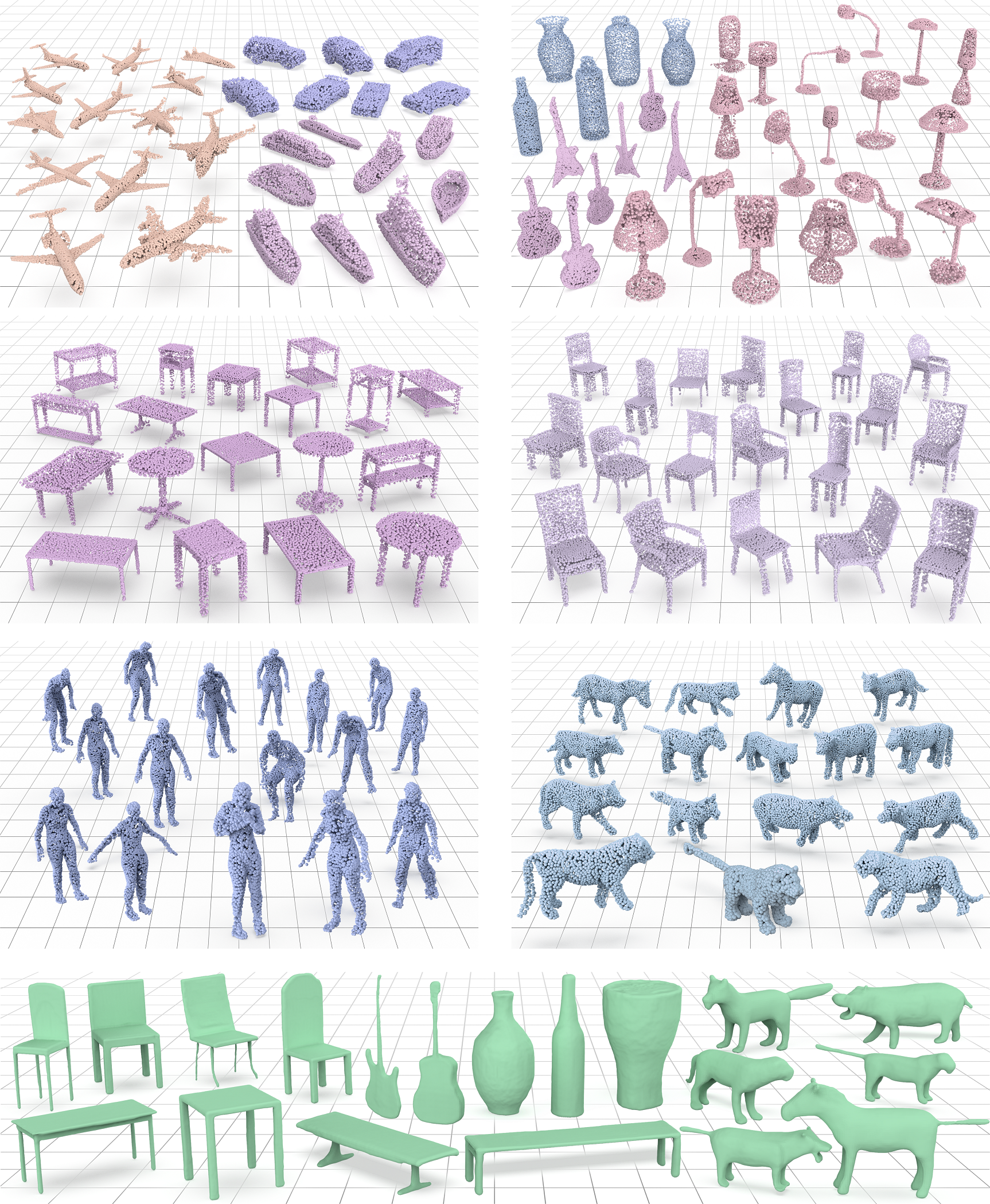}
	\vspace{-1.5mm}
	\caption{Galleries showcasing shapes of various categories generated by SP-GAN, and the rich global structures and fine details exhibited in the shapes.}
	\label{fig:gallery}
	%\vspace{-4mm}
\end{figure*}

%\vspace{.05in}\noindent\textbf{Galleries of generated shapes.} \
\paragraph{Galleries of generated shapes}
Figure~\ref{fig:gallery} showcases varieties of shapes generated by SP-GAN in the form of point clouds, from which we can further reconstruct surfaces using~\cite{Liu2021MLS} (see bottom row).
Besides the commonly-used ShapeNet dataset~\cite{chang2015shapenet}, we adopted SMPL~\cite{SMPL:2015} (human body shapes) and SMAL~\cite{Zuffi:CVPR:2017} (animal shapes) as our training data.
In detail, on the 3D mesh of each shape, we uniformly sampled $N$ points ($N=2,048$ by default) and normalized the points to fit a unit ball.
Like the existing generative models~\cite{achlioptas2018learning,shu20193d,hui2020progressive}, we trained our network on each category separately.
Note that we regard different kinds of animals as individual category.
During the generation phase, we randomly sample a latent code from a standard Gaussian distribution and take it to SP-GAN to generate the associated shape (point cloud).
As shown in Figure~\ref{fig:gallery}, our generated shapes exhibit fine details with less noise and also cover a rich variety of global and local structures.

\begin{figure}[!t]
	\centering
	\includegraphics[width=0.95\linewidth]{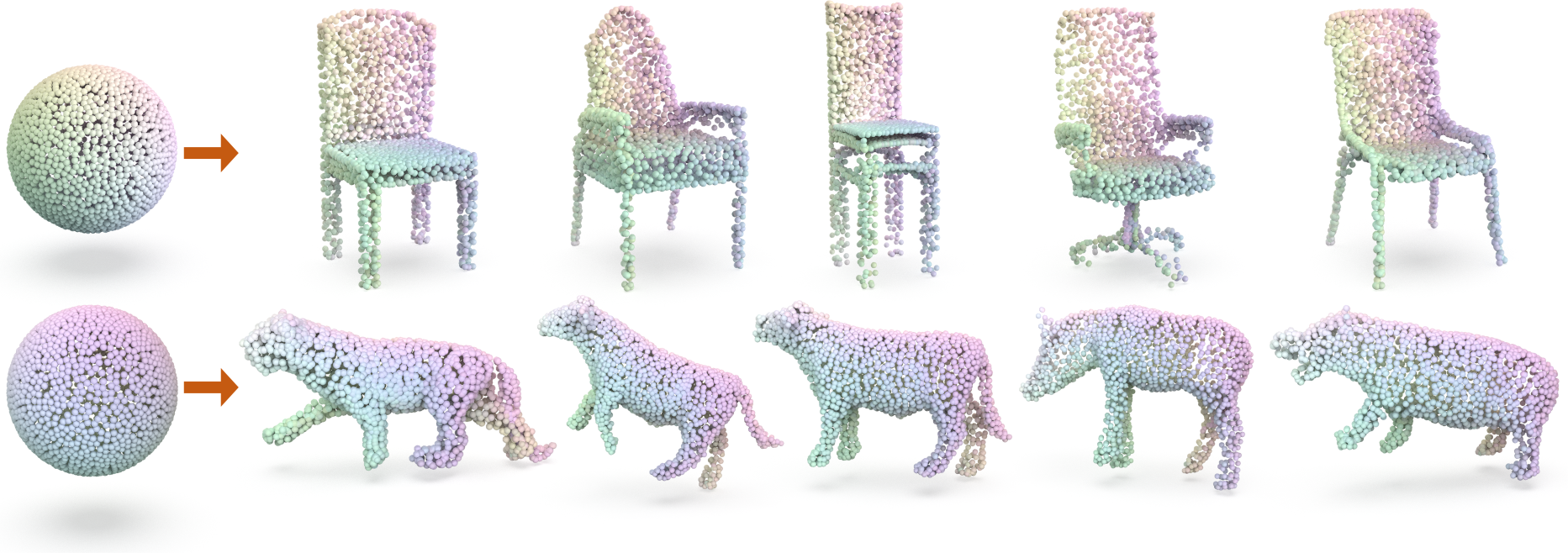}
	\vspace{-4mm}
    \caption{Visualizing the dense correspondence between the sphere proxy and the generated shapes.
	Note that same color is assigned to associated points on the sphere proxy and on the generated shapes.}
	\label{fig:correspondence}
    \vspace{-2mm}
\end{figure}

\begin{figure}[!t]
	\centering
	\includegraphics[width=0.99\linewidth]{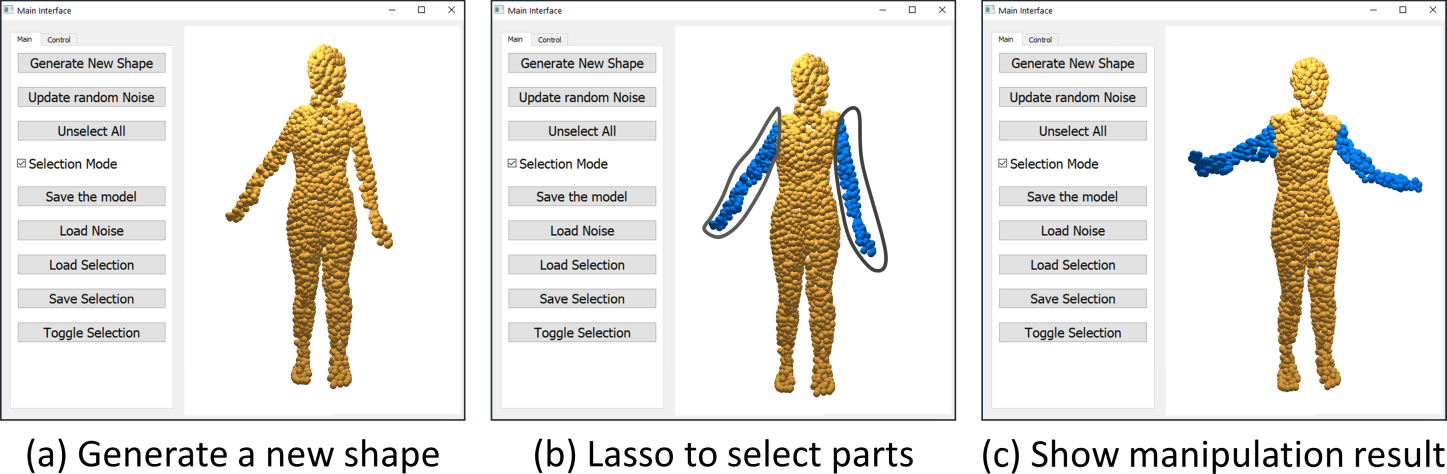}
	\vspace{-2mm}
	\caption{Our interface for interactive shape generation and manipulation.
	Given a generated shape (a), one can use a lasso tool to select specific parts in the shape (b) and preview the manipulation result (c) in real-time.}
	\label{fig:interface}
	\vspace{-3mm}
\end{figure}

%%%%%%%%%%%%%%%%%%%%%%%%%%%%%%%%%%%%%%%%%%%%%%%%%%%%%%%%%%%%%%%

%\vspace{.05in}\noindent\textbf{Visualization of the dense correspondence.} \
\paragraph{Visualization of the dense correspondence}
To visualize the learned implicit dense correspondence between the points on the sphere proxy $\mathcal{S}$ and the points in the generated shapes, we use same color to render associated points on the sphere and on the generated shapes; see Figure~\ref{fig:correspondence} for examples.
Checking the point colors across sphere and generated shapes, we can see that different regions of the sphere correspond to different local parts in the shapes.
For example, the lower left part of the sphere corresponds to the legs of the chairs.
Also, in each category, the same semantic part of different objects,~\eg, the back of all chairs and the legs of all animals, receive similar colors.
Particularly, as shown in the bottom row, these animals have different poses, yet our model can successfully establish a dense correspondence between the semantically-associated parts.
This correspondence property is crucial to enable various structure-aware shape manipulations, as we shall show soon.

%%%%%%%%%%%%%%%%%%%%%%%%%%%%%%%%%%%%%%%%%%%%%%%%%%%%%%%%%%%%%%%

\begin{figure}[!t]
	\centering
	\includegraphics[width=0.99\linewidth]{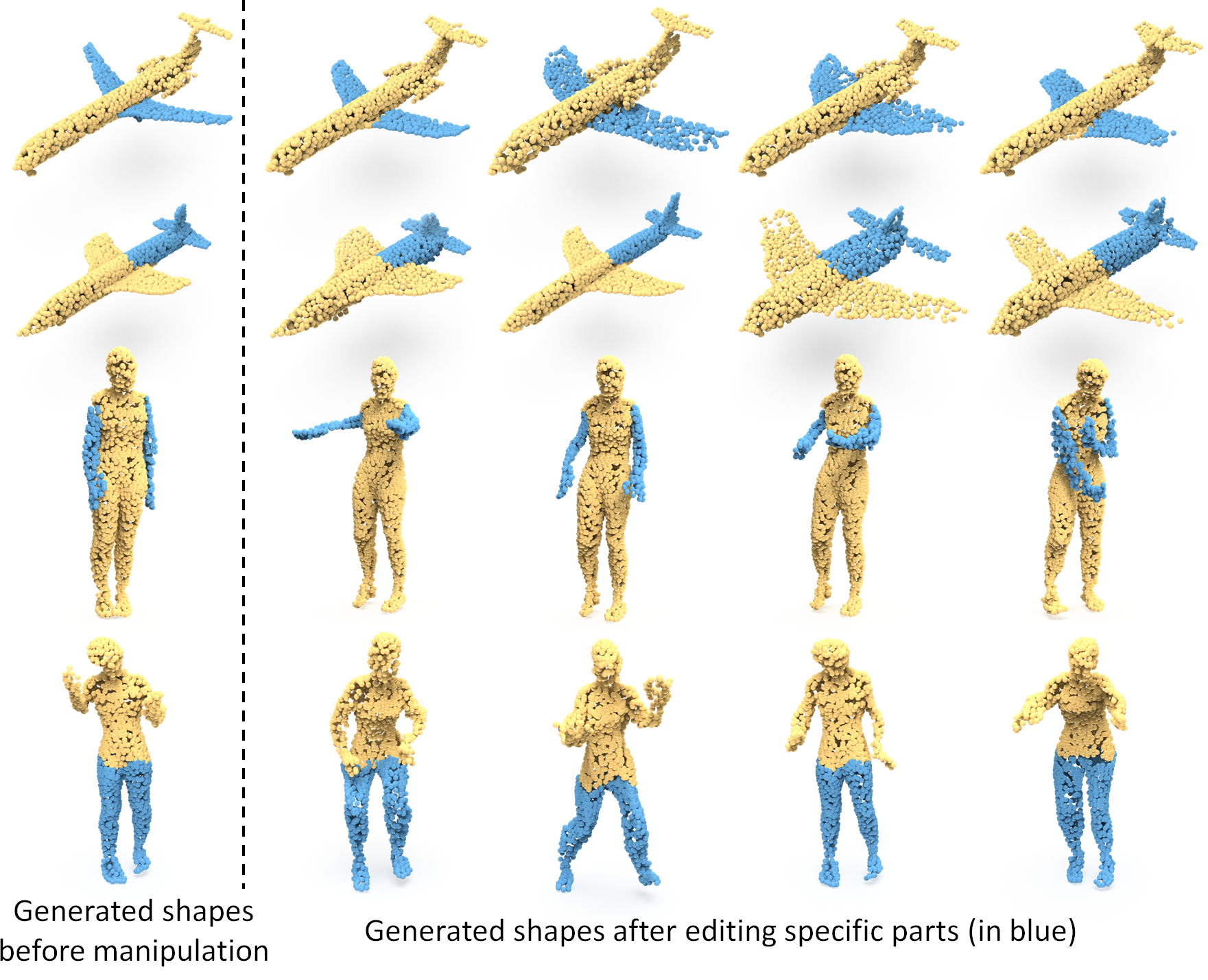}
	\vspace*{-3mm}
	\caption{Part editing examples.
	Given a generated shape (left), we can select specific parts (blue) and modify their associated latent codes to produce new shapes with different styles (right).}
	\label{fig:part_edit}
	\vspace*{-1mm}
\end{figure}

\begin{figure}[!t]
	\centering
	\includegraphics[width=0.99\linewidth]{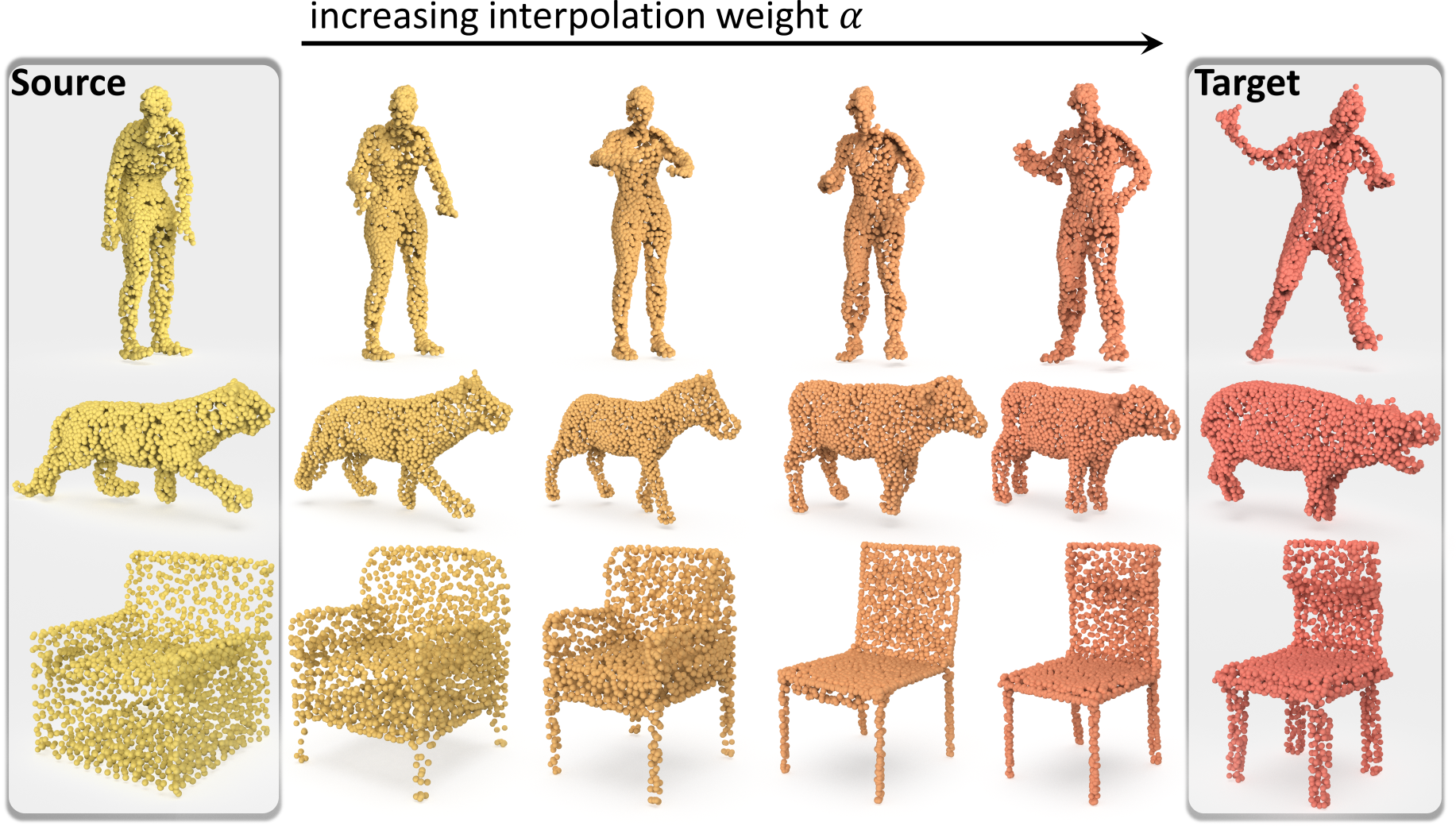}
	\vspace*{-2mm}
	\caption{Shape-wise interpolation between the source (left) and target (right-most) in each row. Note the smooth transition from left to right, including the two human bodies with different poses (top row).
	Also, the generated intermediate shapes are high-quality with fine details and little noise.}
	\label{fig:ShapeInte}
	\vspace*{-2mm}
\end{figure}

\begin{figure}[!t]
	\centering
	\includegraphics[width=0.99\linewidth]{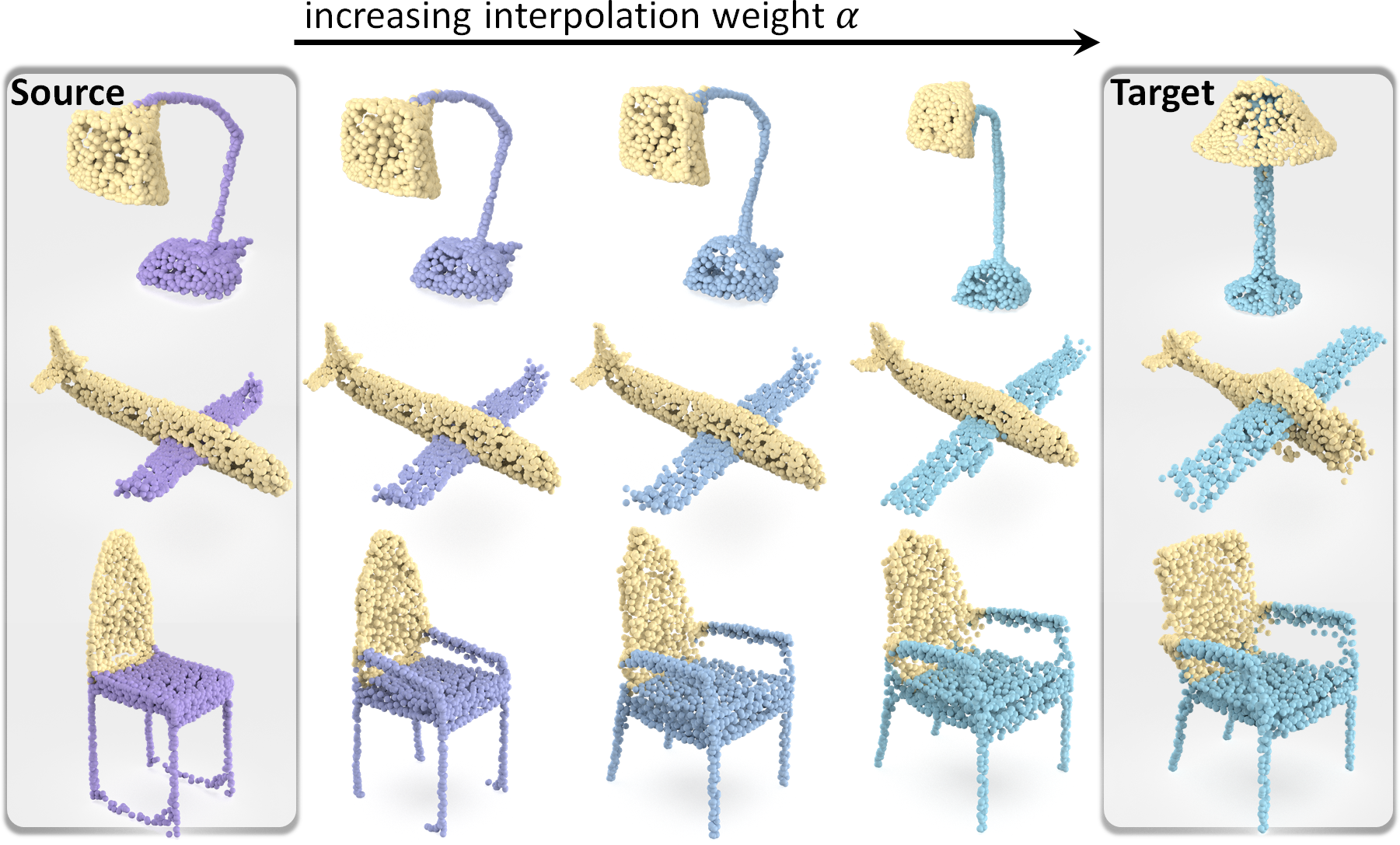}
	\vspace*{-2mm}
	\caption{Interpolating only the latent code associated with the user-selected parts (non-yellow) between the source (left) and target (right) in each row.
	Note the smooth and part-based transition, as well as the quality of the generated point samples in the interpolated shapes.}
	\label{fig:partInte}
	\vspace*{-2mm}
\end{figure}

\subsection{Single-shape Part Editing}
\label{subsec:edit}
With a dense correspondence established between the sphere proxy and each generated shape, we can easily locate ``local'' latent code associated with specific parts in a shape.
As illustrated in Figure~\ref{fig:manipulate_pipeline}(a), we can perform part editing by re-sampling a new random latent code to replace the existing latent code associated with each specific part in the shape; by then, we can generate a new shape that is similar to the original one but with a different local part.

To facilitate part selection, we built the interactive interface shown in Figure~\ref{fig:interface}, in which one can use a lasso tool to interactively select specific parts in a generated shape.
The blue points in (b) indicate the lasso selection.
Then, one can click a button to regenerate the shape with a real-time preview (c).
Please watch our supplemental video.
Figure~\ref{fig:part_edit} shows more part-zediting examples.
On the left of each row is an initial generated shape without any manipulation.
By selecting specific parts (blue) in the shape, we can randomly modify the associated latent codes and produce the new shapes shown on the right.
We can observe that, after modifying a portion of all the latent codes, the generator will synthesize new shapes with different local styles on the user-marked blue parts, while other object parts (yellow) only exhibit small changes for compatibility with the newly-modified parts.

\subsection{Shape Interpolation}
\label{subsec:interpolation}

%\vspace{.05in}\noindent\textbf{Shape-wise interpolation.} \
\paragraph{Shape-wise interpolation}
Like existing models~\cite{achlioptas2018learning,yang2019pointflow,shu20193d}, we can also linearly interpolate between shapes of different appearance using SP-GAN.
Specifically, given two different generated shapes $\mathcal{P}_a$ (source) and $\mathcal{P}_b$ (target) with their associated latent codes $\mathbf{z}_a$ and $\mathbf{z}_b$, respectively, the interpolated shape $\mathcal{P}_c$ can be easily generated by feeding the interpolated latent code $\mathbf{z}_c=(1-\alpha) \cdot \mathbf{z}_a + \alpha \cdot \mathbf{z}_b$ into our generator, where $\alpha$ is the interpolation weight ranged from zero to one.

From the interpolation results shown in Figure~\ref{fig:ShapeInte}, we can observe that as $\alpha$ increases, SP-GAN enables a smooth transition from the source to the target.
See particularly the top row, the source and target reprensent the same human body of two different poses; the interpolation is able to produce rather smooth transitions between the two poses.
Distinctively, just like the source and target, the intermediate (generated) shapes also contain fine details with little noise; this is very challenging to achieve, as many existing works easily introduce noise to the interpolated point clouds.

%\vspace{.05in}\noindent\textbf{Part-wise interpolation.} \
\paragraph{Part-wise interpolation}
As illustrated in Figure~\ref{fig:manipulate_pipeline}(b), thanks to the dense correspondence established across the generated shapes, we can interpolate the latent code associated with specific part in a shape, instead of simply interpolating the whole shape.

Figure~\ref{fig:partInte} shows three sets of part-wise interpolation results, where we interpolate only the latent codes associated with the user-selected points (non-yellow) between the source and target shapes.
Here, we render the user-selected points using a color gradient from purple to blue to reveal the interpolation progress.
From the results shown in the figure, we can observe that the interpolation happens mainly on the user-selected points, yet the remaining points (yellow) may exhibit small changes for the integrity of the generated shape.
For example, in the first row, we select the bracket and base of the source and target lamps for interpolation, so the lamp's shape does not change significantly, while the style of the bracket and base gradually morphs from that of the source to the target.

\begin{figure}[!t]
	\centering
	\includegraphics[width=0.99\linewidth]{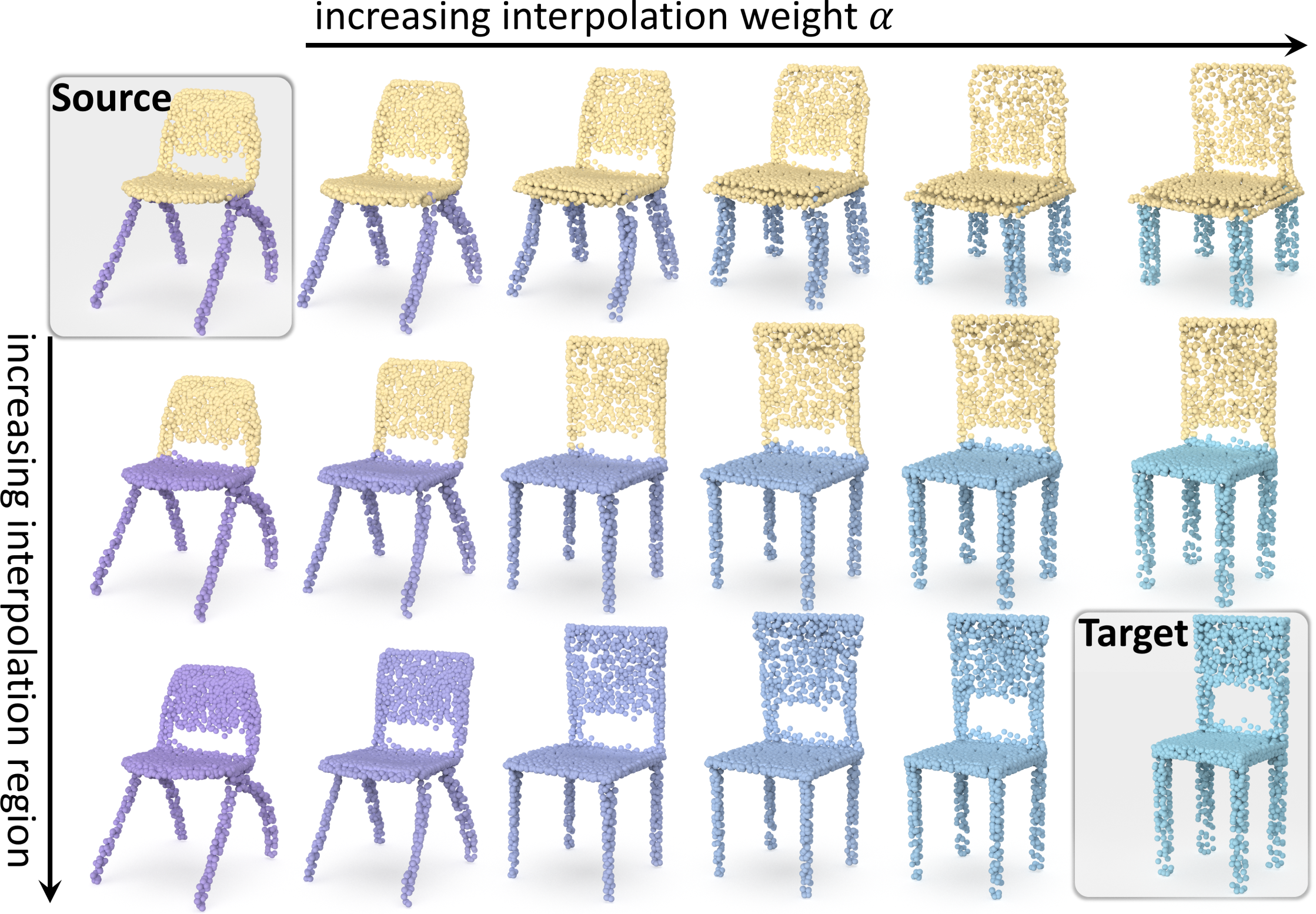}
	\vspace*{-2mm}
	\caption{A two-dimensional part-wise interpolation example.
	Along each row (left to right), we interpolate the user-selected part, whereas along each column (top to bottom), we gradually enlarge the user-selected region.
	From the results shown above, we can see the smooth transition and the quality of the interpolated shapes produced by SP-GAN.}
	\label{fig:part_shape_Interpolation}
	\vspace*{-2mm}
\end{figure}

Figure~\ref{fig:part_shape_Interpolation} further shows a two-dimensional part-wise interpolation example.
Along each row, we fix the part being interpolated (non-yellow) and gradually increase the interpolation weight $\alpha$.
Taking the first row as an example, we only selected the legs for interpolation.
As $\alpha$ increases, the legs of the source (top-left) gradually becomes more and more similar to the legs of the target (bottom-right); see also the color gradient from purple to blue.
On the other hand, along each column (top to bottom), we fix $\alpha$ but gradually enlarge the selected region.
Taking the right-most column as an example, in which $\alpha=1$,
we can observe that as the selected region (blue) increases, the source gradually deforms further to become the target.
Again, we would like to highlight that shapes generated by SP-GAN have fine details and little noise, including also the interpolated ones; see again Figures~\ref{fig:partInte} and~\ref{fig:part_shape_Interpolation}.

%%%%%%%%%%%%%%%%%%%%%%%%%%%%%%%%%%%%%%%%%%%%%%%%%%%%%%%%%%%%%%%

%%%%%%%%%%%%%%%%%%%%%%%%%%%%%%%%%%%%%%%%%%%%%%%%%%%%%%%%%%%%%%%

\subsection{Multi-shape Part Composition}
\label{subsec:multi_edit}
The implicit dense correspondence across multiple generated shapes also facilitates the composition of parts from multiple shapes.
Specifically, given different semantic parts (\eg, chair's back, legs,~\etc) and their associated latent codes from different shapes, we can pack a new prior latent matrix with $\mathcal{S}$ and feed the matrix to our generator to produce new shapes.
In this way, the synthesized shape would integrate different local styles from the parent shapes.

\begin{figure}[!t]
	\centering
	\includegraphics[width=0.95\linewidth]{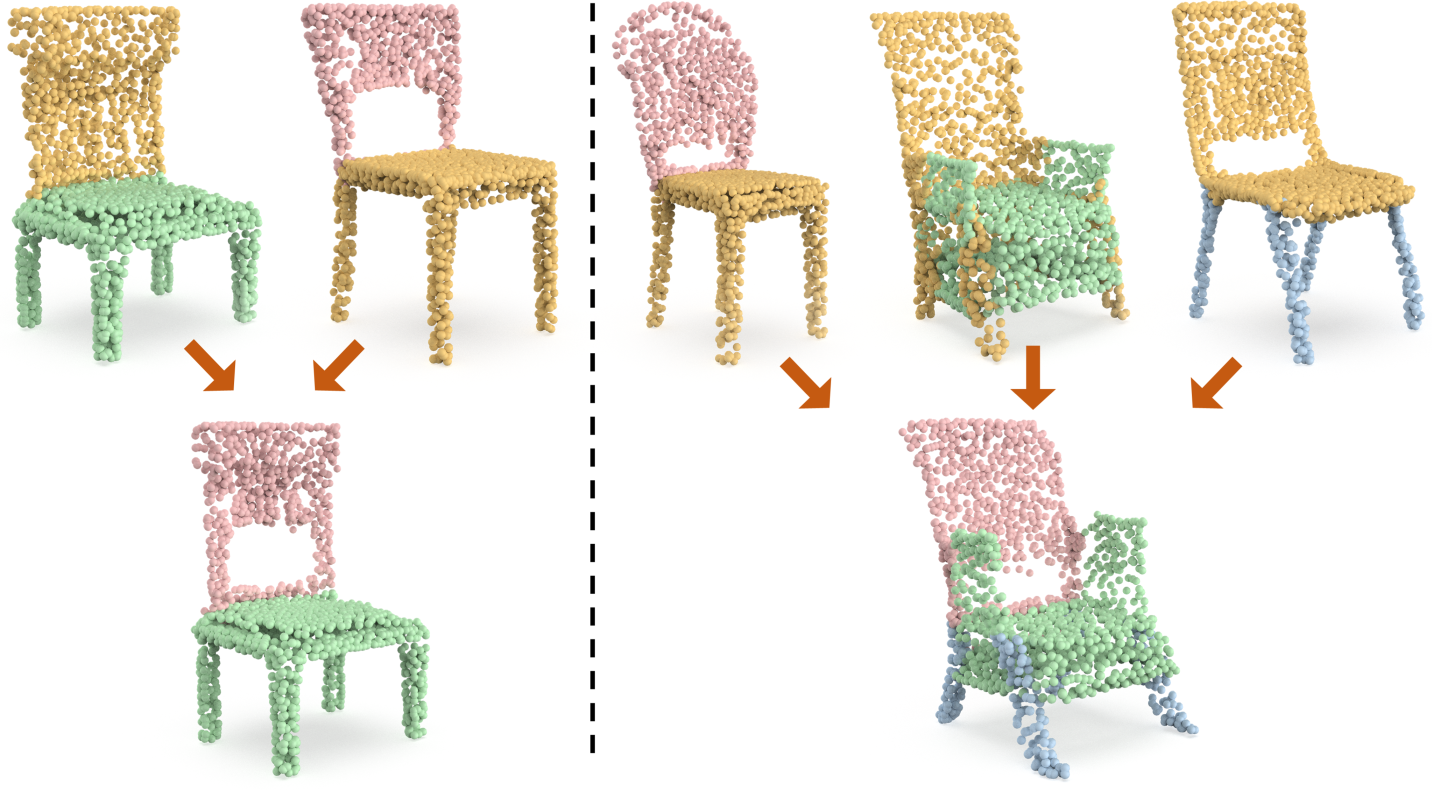}
	\vspace*{-3.5mm}
	\caption{Two multi-shape part composition examples (left and right), in which the non-yellow parts in the top row are selected in the composition.}
	\label{fig:composition}
	\vspace{-2.5mm}
\end{figure}

Figure~\ref{fig:composition} presents two examples, where the non-yellow parts in the shapes (top row) are selected for composition.
We can observe that, for the two re-generated shapes (bottom row), their local parts generally inherit the styles of the parent shapes.

%%%%%%%%%%%%%%%%%%%%%%%%%%%%%%%%%%%%%%%%%%%%%%%%%%%%%%%%%%%%%%%
\begin{figure}[t]
	\centering
	\includegraphics[width=0.95\linewidth]{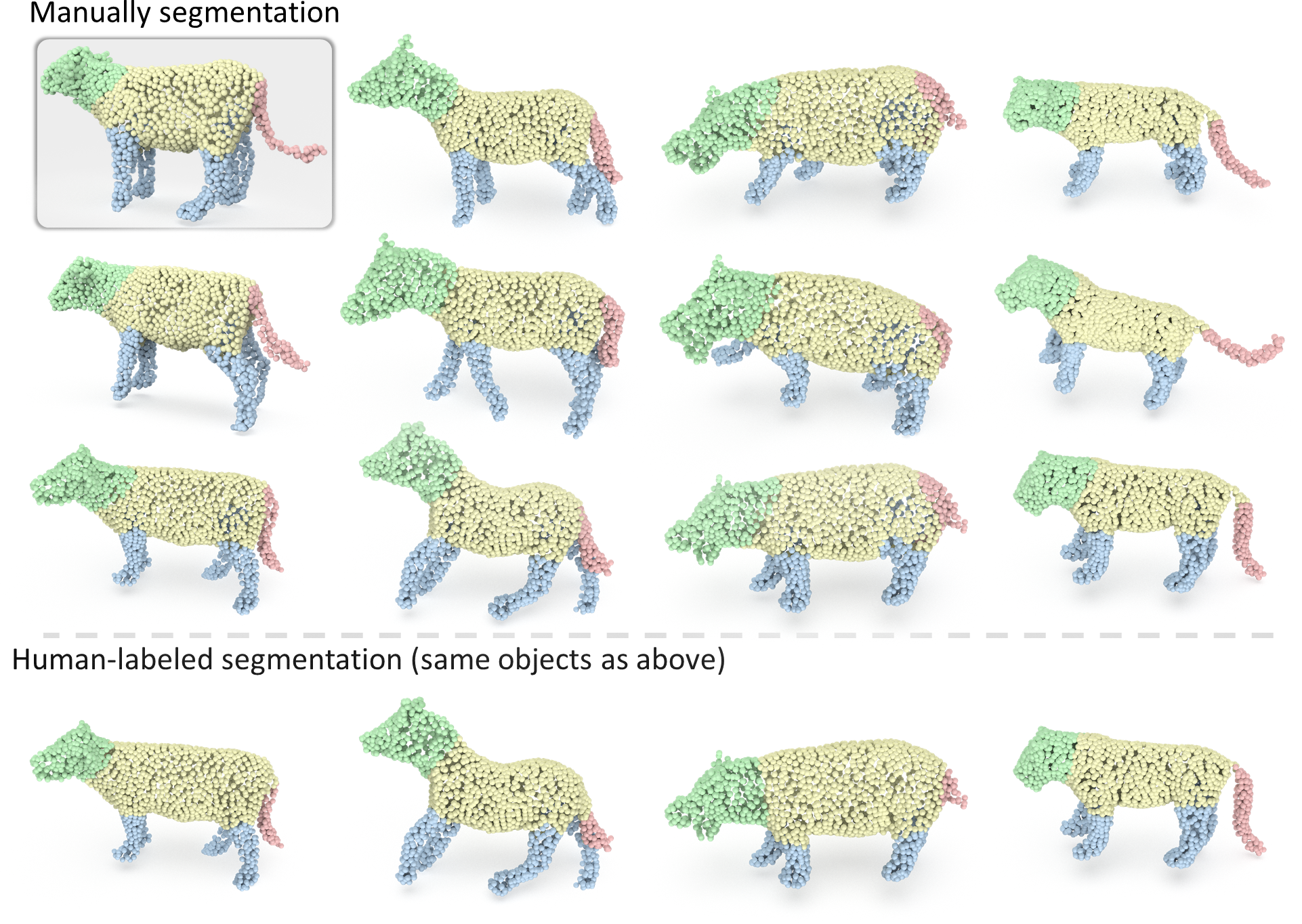}
	\vspace*{-4mm}
	\caption{Top: co-segmentation results relying on the dense correspondence across different generated shapes, where only the top-left object (marked over a gray background) is manually segmented and the others are automatically co-segmented. Note that each column represents a different kind of animal. Bottom row: the associated human-labeled results.}
	\label{fig:coseg}
	\vspace{-2.5mm}
\end{figure}

\subsection{Part Co-Segmentation}
\label{subsec:unsupervised}

Another benefit brought by the implicit dense correspondence is that SP-GAN enables part co-segmentation on a set of 3D shapes,~\ie, simultaneous segmentation of shapes into semantically-consistent parts.
To do so, we only need to manually segment one of the shapes in the set, then the point-wise labels of the remaining shapes could be simultaneously obtained through the dense correspondence.

Figure~\ref{fig:coseg} showcases an interesting co-segmentation example, in which only the top-left shape was manually segmented and all other shapes in the figure are {\em automatically\/} co-segmented.
We can observe that, the learned implicit correspondence produced by SP-GAN can successfully facilitate part co-segmentation across shapes, even for shapes with different poses and local part orientations,~\eg, see the tails of the animals in the figure.
Particularly, different columns actually feature different types of animals, yet our method can still achieve high robustness in co-segmenting these shapes.

Since generative tasks have no ground truths, we cannot directly evaluate the dense correspondence between the generated results.  So, we resort to comparing our shape co-segmentation results with human-labeled results.  Specifically, we manually label parts in 100 generated shapes of SMAL and calculate the mIoU between the co-segmentation results and human labels.  The results are very positive (87.7\%).  Also, we conducted a user study with 10 participants (who are graduate students aged 23 to 28), to compare the 1-shot co-segmented sample and human-labeled sample; the preference statistics of our result is 48.5\%, which is close to a random selection. Figure~\ref{fig:coseg} (bottom) shows some of the human-labeled results.

%%%end of section

\section{Evaluation and Discussion}
\label{sec:evaluation}

\begin{table*}[t]
\centering
\caption{Quantitative comparison of the generated shapes produced by SP-GAN and five state-of-the-art methods,~\ie, r-GAN~\cite{achlioptas2018learning}, tree-GAN~\cite{shu20193d}, PointFlow~\cite{yang2019pointflow}, PDGN~\cite{hui2020progressive}, and ShapeGF~\cite{cai2020learning}.
We follow the same settings to conduct this experiment as in the state-of-the-art methods.
From the table, we can see that the generated shapes produced by our method have the best quality (lowest MMD, largest COV, and lowest FPD) for both the Airplane and Chair datasets.
Also, our method has the lowest complexity (fewest model parameters) and the highest training efficiency (shortest model training time).
The unit of MMD is $10^{-3}$.}
\label{tab:quanComparison}
\vspace*{-3mm}
\resizebox{0.8\linewidth}{!}{
	\begin{tabular}{C{1.5cm}|C{1.5cm}|C{1.5cm}C{1.5cm}C{1.5cm}C{1.5cm}C{1.5cm}C{1.5cm}}
		\toprule[1pt]
		\multirow{2}*{Categories} & \multirow{2}*{Metrics} & \multicolumn{6}{c}{Methods} \\ \cline{3-8}
		& & r-GAN & tree-GAN & PointFlow & PDGN & ShapeGF & SP-GAN \\ \hline \hline
		\multirow{3}*{Airplane} & MMD($\downarrow$) &3.81 & 4.32 & 3.68 & 3.27 & 3.72 & \textbf{1.95} \\
		& COV(\%,$\uparrow$) & 42.17 & 39.37 & 44.98 & 45.68 & 46.79 & \textbf{50.50} \\
		& FPD($\downarrow$) & 3.54 & 2.98 & 2.01 & 1.84 & 1.61 & \textbf{0.96} \\ \hline
		\multirow{3}*{Chair} & MMD($\downarrow$) & 18.18 & 16.14 & 15.02 & 15.56 & 14.81 & \textbf{8.24} \\
		& COV(\%,$\uparrow$) & 38.52 & 39.37 & 44.98 & 45.89 & 47.79 & \textbf{52.10} \\
		& FPD($\downarrow$) & 5.28 & 4.44 & 3.83 & 3.77 & 3.52 & \textbf{2.13} \\ \hline
		\multicolumn{2}{c|}{Parameter size (M)} & 7.22 & 40.69 & 1.61 & 12.01 & 5.23 & \textbf{1.01} \\ \hline
		\multicolumn{2}{c|}{Training time (Days)} & 0.9 & 4.1 & 2.5 & 2.3 & 1.2 & \textbf{0.5} \\
		\bottomrule[1pt]
\end{tabular}}
\end{table*}

\begin{figure}[!t]
	\centering
	\includegraphics[width=0.96\linewidth]{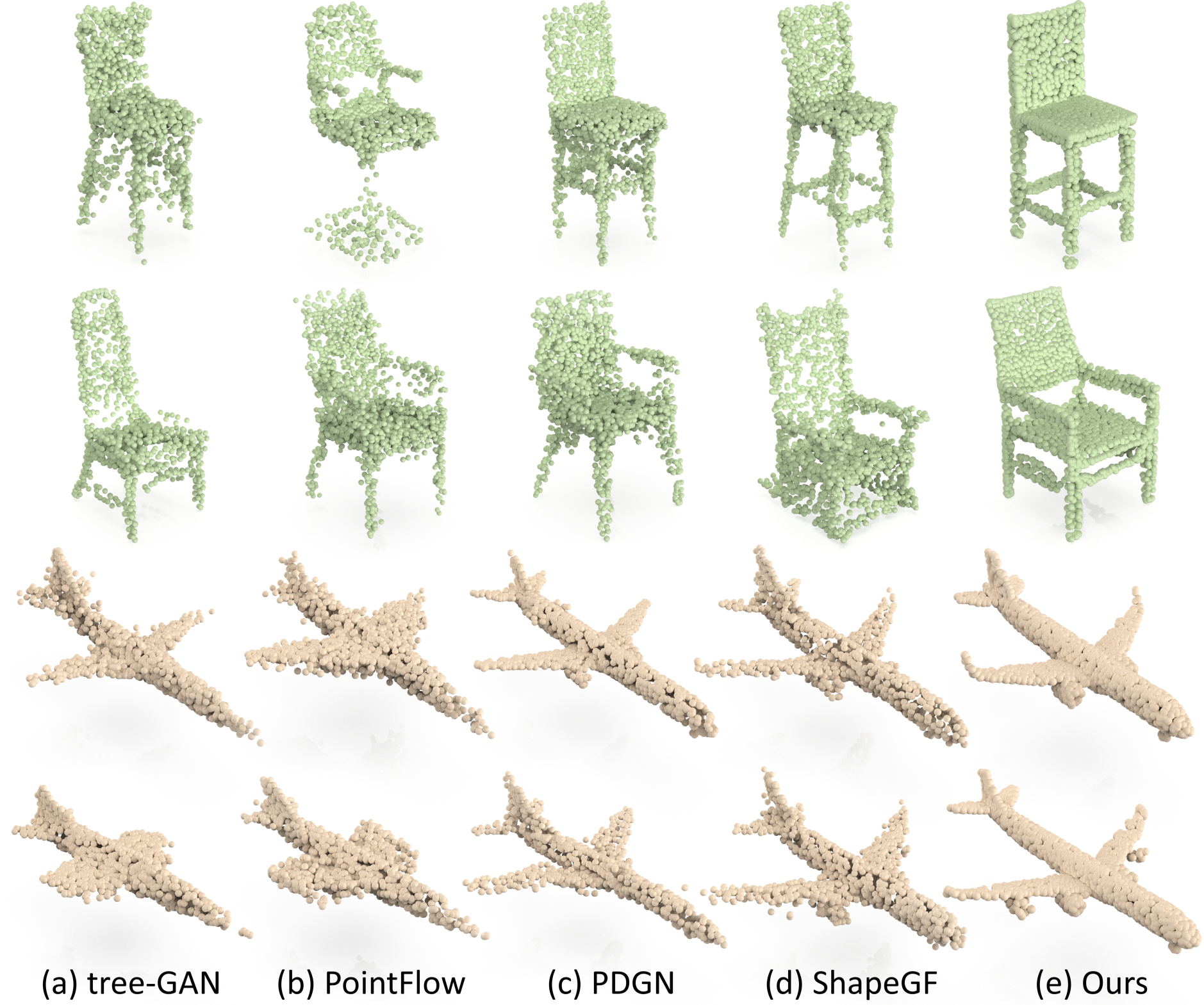}
	\vspace{-2mm}
	\caption{Visual comparisons with state-of-the-art methods.
	Clearly, the point clouds generated by our method (e) exhibit more fine details and less noise, while other methods (a)-(d) tend to generate noisy point samples. }
	\label{fig:visComparison}
	\vspace{-3mm}
\end{figure}

%%%%%%%%%%%%%%%%%%%%%%%%%%%%%%%%%%%%%%%%%%%%%%%%%%%%%

\subsection{Comparing with Other Methods}
\label{subsec:comparison}

We evaluate the shape generation capability of our method against five state-of-the-art generative models, including r-GAN~\cite{achlioptas2018learning}, tree-GAN~\cite{shu20193d}, PointFlow~\cite{yang2019pointflow}, PDGN~\cite{hui2020progressive}, and ShapeGF~\cite{cai2020learning}.
Here, we follow the same experimental setup as the above works to conduct this experiment.
That is, we trained our SP-GAN model also on the Chair (and Airplane) shapes in ShapeNet~\cite{chang2015shapenet}, randomly generated 1,000 shapes for each category, and then evaluated the generated shapes using the same set of metrics as in the previous works (details to be provided below).
Also, for the five state-of-the-art models being compared, we directly employ their publicly-released trained network models to generate shapes.

%\para{Evaluation metrics.} \
\paragraph{Evaluation metrics}
Different from supervised tasks, generation tasks have no explicit ground truths for evaluations.
Hence, we directly follow the evaluation metrics in the existing works~\cite{achlioptas2018learning,shu20193d,hui2020progressive}:
(i) Minimum Matching Distance (MMD) measures how close the set of generated shapes is relative to the shapes in the given 3D repository $\{\mathcal{\hat{P}}\}$, so MMD indicates the fidelity of the generated shapes relative to $\{\mathcal{\hat{P}}\}$.
(ii) Coverage (COV) measures the fraction of shapes in $\{\mathcal{\hat{P}}\}$ that can be matched (as a nearest shape) with at least one generated shape, so COV indicates how well the generated shapes cover the shapes in $\{\mathcal{\hat{P}}\}$;
and
(iii) Fr$\acute{e}$chet Point Cloud Distance (FPD) measures the 2-Wasserstein distance in the feature space between the generated shapes and the shapes in $\{\mathcal{\hat{P}}\}$, where we employ a pre-trained classification model,~\ie, DGCNN~\cite{wang2019dynamic}, for the feature extraction, so FPD indicates the distribution similarity between the generated shapes and $\{\mathcal{\hat{P}}\}$.
For details of these metrics, readers may refer to~\cite{achlioptas2018learning,shu20193d,hui2020progressive}.
Although these metrics are not absolutely precise, they still reflect the quality of the generated shapes in various aspects.
Overall, a good method should have a low MMD, high COV, and low FPD.

\paragraph{Quantitative evaluation.} 
Table~\ref{tab:quanComparison} reports the quantitative comparison results between different methods.
For a fair comparison of training time, all methods (including ours) were run on the same desktop computer with the same GPU, and for the five comparison methods, we used code from their project websites.
From the results shown in the table, we can see that SP-GAN outperforms all the other methods consistently on the three evaluation metrics for both datasets by a large margin.
The low MMD and FPD suggest that our generated shapes have high fidelity compared with the shapes in the 3D repository in both the spatial and feature space, and the high COV suggests that our generated shapes have a good coverage of the shapes in the 3D repository.
Also, our network has the fewest learnable parameters and requires much less time to train.

\paragraph{Qualitative evaluation.} 
Figure~\ref{fig:visComparison} shows some visual comparison results.
Here, we pick a random shape generated by our method and use the Chamfer distance to retrieve the nearest shapes generated by each of the other comparison methods.
From these results, we can see that the point clouds generated by our method (e) clearly exhibit more fine details and less noise, while other methods (a)-(d) tend to generate noisy point samples.

%%%%%%%%%%%%%%%%%%%%%%%%%%%%%%%%%%%%%%%%%%%%%%%%%%%%%

\subsection{Ablation Study}
\label{subsec:ablation}

To evaluate the effectiveness of the major components in our framework, we conducted an ablation study by simplifying our full pipeline for four cases: (i) replace the graph attention module (GAM) with the original EdgeConv~\cite{wang2019dynamic}; (ii) remove the adaptive instance normalization (AdaIN) and directly take the prior latent matrix as the only input (see Figure~\ref{fig:generator}); (iii) remove per-point score (PPS) (see Figure~\ref{fig:discriminator}); and (iv) replace the sphere with a unit cube.
In each case, we re-trained the network and tested the performance on the Chair category.
Table~\ref{tab:ablation} summarizes the results of (i) to (iv), demonstrating that our full pipeline (bottom row) performs the best and removing any component reduces the overall performance, thus revealing that each component contributes.

\begin{table}[!t]
	\caption{Comparing the generation performance of our full pipeline with various simplified cases in the ablation study. The unit of MMD is $10^{-3}$.}
	\centering
    \vspace{-2mm}
	\label{tab:ablation}
	\resizebox{0.95\linewidth}{!}{%
		\begin{tabular}{c|cccc|ccc}
			\toprule[1pt]%\noalign{\smallskip}
			Model & GAM & AdaIN   & PPS & Sphere & MMD($\downarrow$)& COV(\%,$\uparrow$)& FPD($\downarrow$)\\
            \hline
            (i) &  &$\checkmark$  &  $\checkmark$ & $\checkmark$   & 10.38 &  48.24&  2.88\\	
            (ii) & $\checkmark$  &    & $\checkmark$ & $\checkmark$ & 9.54 &  49.72&  2.66 \\	
			(iii) & $\checkmark$  &  $\checkmark$  &   & $\checkmark$&  9.81&  50.63&  2.52  \\
			(iv) & $\checkmark$  & $\checkmark$   & $ \checkmark$ & & 8.69&  51.22&  2.29\\
			\hline
			Full & $\checkmark$ & $\checkmark$ & $\checkmark$ & $\checkmark$ &\textbf{8.24}&\textbf{52.10}&\textbf{2.13}
			\\
			\bottomrule[1pt]
	\end{tabular}}
	%\end{center}
	\vspace*{-1.5mm}
\end{table}

%%%%%%%%%%%%%%%%%%%%%%%%%%%%%%%%%%%%%%%%%%%%%%%%%%%%%

\subsection{Shape Diversity Analysis}
\label{subsec:analysis}

\begin{figure}[!t]
\centering
\includegraphics[width=0.93\linewidth]{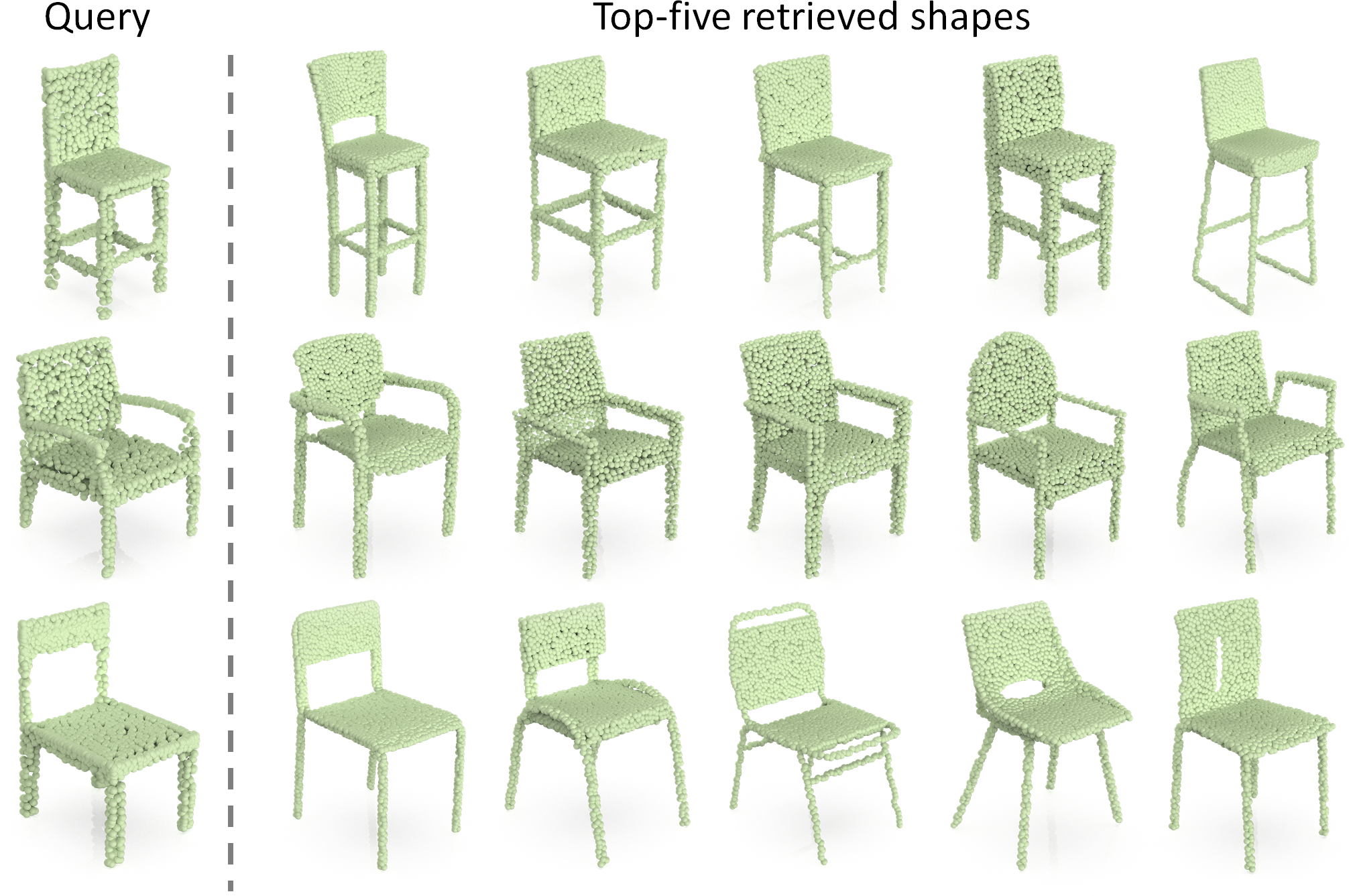}
\vspace{-2mm}
\caption{Shape diversity analysis.
We take shapes (left) randomly produced by our generator to query the shapes in the training set (the given 3D repository).
The top-five most similar shapes for each case are presented on the right, showing that our generated shapes look similar to the retrieved shapes in overall structure, yet they have different local fine details.}
\label{fig:query}
\vspace{-1.5mm}
\end{figure}

Further, to show that our generator is able to synthesize diverse shapes with fine details and it does not simply memorize the latent code of the shapes in the training set $\{\mathcal{\hat{P}}\}$, we conducted a shape retrieval experiment.
Specifically, we take random shapes produced by our generator to query the shapes in the training set,~\ie, we find the shapes in training set $\{\mathcal{\hat{P}}\}$ that have the smallest Chamfer distance with each generated shape.
Figure~\ref{fig:query} shows some of the shape retrieval results.
Compared with the top-five similar shapes retrieved from $\{\mathcal{\hat{P}}\}$, we can see that our generated shapes have similar overall structures yet different local fine details.
Particularly, the points in our generated shapes have less noise (compared with those produced by other unsupervised generative models), just like the sampled ones from the training set.

\subsection{Discussion on Limitations}
\label{subsec:limit}

\begin{figure}[!t]
	\centering
	\includegraphics[width=0.99\linewidth]{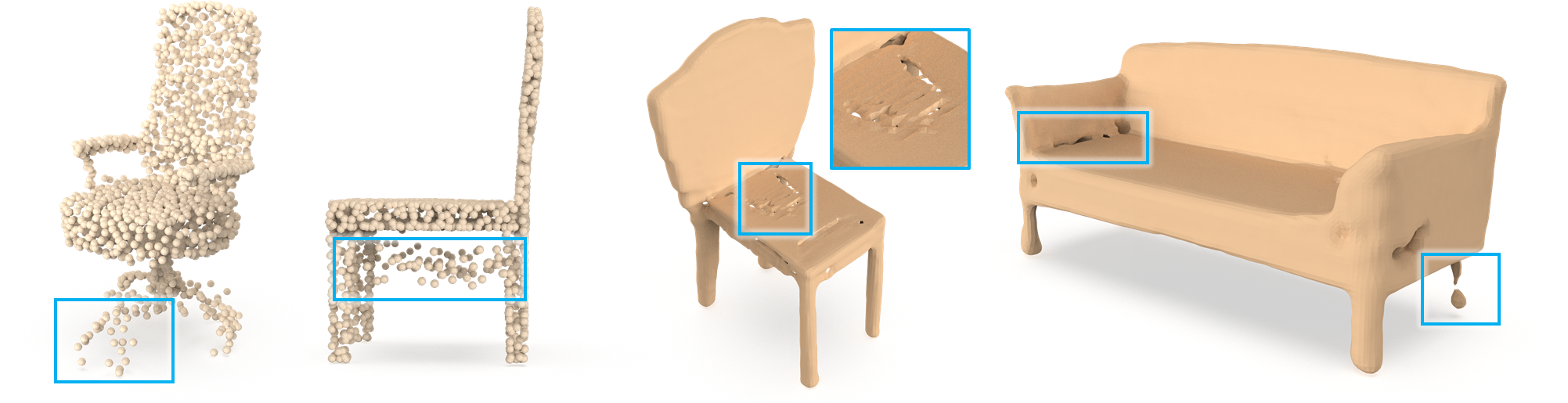}
	\vspace*{-2mm}
	\caption{Failure cases.
	Left: complex and thin structures may be blurry.
	Right: distorted edges and holes could be produced on the reconstructed surface.}
	\label{fig:failures}
	\vspace{-2mm}
\end{figure}

Our approach still has several limitations.
(i) SP-GAN is able to learn in an unsupervised manner but it still requires a large amount of shapes for training.
Hence, for shapes with limited training samples or with complex or thin structures, the generated shapes may still be blurry (or noisy);
see Figure~\ref{fig:failures} (left).
(ii) Though the point cloud representation is flexible for shape generation and manipulation, we can not directly produce surface or topological information and require a post processing to reconstruct the surface. So, distorted edges and holes could be produced on the reconstructed surface; see Figure~\ref{fig:failures} (right).
In the future, we plan to explore point normals in the generation process to enhance the mesh reconstruction.
(iii) Lastly, parts relations are not explicitly or clearly represented in our model, even it embeds an implicit correspondence; looking at the human-labeled samples in Figure~\ref{fig:composition}, we may incorporate certain parts priors into the network to enrich the generation.

%%%%%%%%%%%%%%%%%%%%%%%%%%%%%%%%%%%%%%%%%%%%%%%%%%%%%

%%%end of section 
\section{Conclusion And Future Work}
\label{sec:conclusion}

This work presents SP-GAN, a new generative model for direct generation of 3D shapes represented as point clouds.
By formulating the generator input using the prior latent matrix, we decouple the input into a global prior (sphere points) and a local prior (random latent code), and formulate the generator network with style embedding and adaptive instance normalization to bring local styles from the random latent code into the point features of the sphere points.
This disentanglement scheme articulates the 3D shape generation task as a global shape modeling and a local structure regulation, and enables the generative process to start from a shared global initialization, yet accommodating the spatial variations.

Very importantly, distinctive from existing works on direct point cloud generation, our new design introduces structure controllability into the generative process through the implicit dense correspondence.
So, we can modify or interpolate latent codes in a shape-wise or part-wise manner, and enable various forms of structure-aware shape manipulations that cannot be achieved by previous works on direct point cloud generation.
Both quantitative and qualitative experimental results demonstrate that SP-GAN is able to generate diverse, new, and realistic shapes that exhibit finer details and less noise, beyond the generation capability of the previous works.

In the future, we plan to extend our current design by considering different forms of user feedbacks and to enable user-guided automatic shape generation.
Also, we would like to explore the possibility of generating 3D shapes represented in other forms, such as polygonal meshes and implicit surfaces.
%%%end of section

%1130140925535334280

\begin{acks}
The authors would like to thank the anonymous reviewers for their valuable comments.
This work is supported by Research Grants Council of the Hong Kong Special Administrative Region (Project No. CUHK 14206320 \& 14201717 \& 14201918).
\end{acks}
%the Hong Kong Centre for Logistics Robotics, and

% DO NOT INCLUDE ACKNOWLEDGMENTS IN AN ANONYMOUS SUBMISSION TO SIGGRAPH 2019
%\begin{acks}
%
%The authors would like to thank Dr. Maura Turolla of Telecom
%Italia for providing specifications about the application scenario.
%
%The work is supported by the \grantsponsor{GS501100001809}{National
%  Natural Science Foundation of
%  China}{http://dx.doi.org/10.13039/501100001809} under Grant
%No.:~\grantnum{GS501100001809}{61273304\_a}
%and~\grantnum[http://www.nnsf.cn/youngscientists]{GS501100001809}{Young
%  Scientists' Support Program}.
%
%
%\end{acks}

% Bibliography
\bibliographystyle{ACM-Reference-Format}
\bibliography{bibliography}


\begin{thebibliography}{50}

%%% the Web address.

\ifx \showCODEN    \undefined \def \showCODEN     #1{\unskip}     \fi
\ifx \showDOI      \undefined \def \showDOI       #1{#1}\fi
\ifx \showISBNx    \undefined \def \showISBNx     #1{\unskip}     \fi
\ifx \showISBNxiii \undefined \def \showISBNxiii  #1{\unskip}     \fi
\ifx \showISSN     \undefined \def \showISSN      #1{\unskip}     \fi
\ifx \showLCCN     \undefined \def \showLCCN      #1{\unskip}     \fi
\ifx \shownote     \undefined \def \shownote      #1{#1}          \fi
\ifx \showarticletitle \undefined \def \showarticletitle #1{#1}   \fi
\ifx \showURL      \undefined \def \showURL       {\relax}        \fi
% invisible to TeX
\providecommand\bibfield[2]{#2}
\providecommand\bibinfo[2]{#2}
\providecommand\natexlab[1]{#1}
\providecommand\showeprint[2][]{arXiv:#2}

\bibitem[\protect\citeauthoryear{Aberman, Katzir, Zhou, Luo, Sharf, Greif,
  Chen, and Cohen-Or}{Aberman et~al\mbox{.}}{2017}]%
        {aberman2017dip}
\bibfield{author}{\bibinfo{person}{Kfir Aberman}, \bibinfo{person}{Oren
  Katzir}, \bibinfo{person}{Qiang Zhou}, \bibinfo{person}{Zegang Luo},
  \bibinfo{person}{Andrei Sharf}, \bibinfo{person}{Chen Greif},
  \bibinfo{person}{Baoquan Chen}, {and} \bibinfo{person}{Daniel Cohen-Or}.}
  \bibinfo{year}{2017}\natexlab{}.
\newblock \showarticletitle{Dip transform for {3D} shape reconstruction}.
\newblock \bibinfo{journal}{\emph{ACM Transactions on Graphics (SIGGRAPH)}}
  \bibinfo{volume}{36}, \bibinfo{number}{4} (\bibinfo{year}{2017}),
  \bibinfo{pages}{79:1--79:11}.
\newblock


\bibitem[\protect\citeauthoryear{Achlioptas, Diamanti, Mitliagkas, and
  Guibas}{Achlioptas et~al\mbox{.}}{2018}]%
        {achlioptas2018learning}
\bibfield{author}{\bibinfo{person}{Panos Achlioptas}, \bibinfo{person}{Olga
  Diamanti}, \bibinfo{person}{Ioannis Mitliagkas}, {and}
  \bibinfo{person}{Leonidas~J. Guibas}.} \bibinfo{year}{2018}\natexlab{}.
\newblock \showarticletitle{Learning representations and generative models for
  {3D} point clouds}. In \bibinfo{booktitle}{\emph{Proceedings of International
  Conference on Machine Learning (ICML)}}. \bibinfo{pages}{40--49}.
\newblock


\bibitem[\protect\citeauthoryear{Arshad and Beksi}{Arshad and Beksi}{2020}]%
        {arshad2020progressive}
\bibfield{author}{\bibinfo{person}{Mohammad~Samiul Arshad} {and}
  \bibinfo{person}{William~J. Beksi}.} \bibinfo{year}{2020}\natexlab{}.
\newblock \showarticletitle{A progressive conditional generative adversarial
  network for generating dense and colored {3D} point clouds}. In
  \bibinfo{booktitle}{\emph{International Conference on 3D Vision (3DV)}}.
\newblock


\bibitem[\protect\citeauthoryear{Cai, Yang, Averbuch-Elor, Hao, Belongie,
  Snavely, and Hariharan}{Cai et~al\mbox{.}}{2020}]%
        {cai2020learning}
\bibfield{author}{\bibinfo{person}{Ruojin Cai}, \bibinfo{person}{Guandao Yang},
  \bibinfo{person}{Hadar Averbuch-Elor}, \bibinfo{person}{Zekun Hao},
  \bibinfo{person}{Serge Belongie}, \bibinfo{person}{Noah Snavely}, {and}
  \bibinfo{person}{Bharath Hariharan}.} \bibinfo{year}{2020}\natexlab{}.
\newblock \showarticletitle{Learning gradient fields for shape generation}. In
  \bibinfo{booktitle}{\emph{European Conference on Computer Vision (ECCV)}}.
\newblock


\bibitem[\protect\citeauthoryear{Chang, Funkhouser, Guibas, Hanrahan, Huang,
  Li, Savarese, Savva, Song, Su, et~al\mbox{.}}{Chang et~al\mbox{.}}{2015}]%
        {chang2015shapenet}
\bibfield{author}{\bibinfo{person}{Angel~X. Chang}, \bibinfo{person}{Thomas
  Funkhouser}, \bibinfo{person}{Leonidas~J. Guibas}, \bibinfo{person}{Pat
  Hanrahan}, \bibinfo{person}{Qixing Huang}, \bibinfo{person}{Zimo Li},
  \bibinfo{person}{Silvio Savarese}, \bibinfo{person}{Manolis Savva},
  \bibinfo{person}{Shuran Song}, \bibinfo{person}{Hao Su}, {et~al\mbox{.}}}
  \bibinfo{year}{2015}\natexlab{}.
\newblock \showarticletitle{{ShapeNet}: An information-rich {3D} model
  repository}.
\newblock \bibinfo{journal}{\emph{arXiv preprint arXiv:1512.03012}}
  (\bibinfo{year}{2015}).
\newblock


\bibitem[\protect\citeauthoryear{Chen and Zhang}{Chen and Zhang}{2019}]%
        {chen2019learning}
\bibfield{author}{\bibinfo{person}{Zhiqin Chen} {and} \bibinfo{person}{Hao
  Zhang}.} \bibinfo{year}{2019}\natexlab{}.
\newblock \showarticletitle{Learning implicit fields for generative shape
  modeling}. In \bibinfo{booktitle}{\emph{IEEE Conference on Computer Vision
  and Pattern Recognition (CVPR)}}. \bibinfo{pages}{5939--5948}.
\newblock


\bibitem[\protect\citeauthoryear{Deng, Yang, and Tong}{Deng
  et~al\mbox{.}}{2021}]%
        {deng2020deformed}
\bibfield{author}{\bibinfo{person}{Yu Deng}, \bibinfo{person}{Jiaolong Yang},
  {and} \bibinfo{person}{Xin Tong}.} \bibinfo{year}{2021}\natexlab{}.
\newblock \showarticletitle{Deformed Implicit Field: Modeling {3D} Shapes with
  Learned Dense Correspondence}. In \bibinfo{booktitle}{\emph{IEEE Conference
  on Computer Vision and Pattern Recognition (CVPR)}}.
\newblock


\bibitem[\protect\citeauthoryear{Dinh, Sohl-Dickstein, and Bengio}{Dinh
  et~al\mbox{.}}{2016}]%
        {dinh2016density}
\bibfield{author}{\bibinfo{person}{Laurent Dinh}, \bibinfo{person}{Jascha
  Sohl-Dickstein}, {and} \bibinfo{person}{Samy Bengio}.}
  \bibinfo{year}{2016}\natexlab{}.
\newblock \showarticletitle{Density estimation using real {NVP}}. In
  \bibinfo{booktitle}{\emph{International Conference on Learning
  Representations (ICLR)}}.
\newblock


\bibitem[\protect\citeauthoryear{Dubrovina, Xia, Achlioptas, Shalah, Groscot,
  and Guibas}{Dubrovina et~al\mbox{.}}{2019}]%
        {dubrovina2019composite}
\bibfield{author}{\bibinfo{person}{Anastasia Dubrovina}, \bibinfo{person}{Fei
  Xia}, \bibinfo{person}{Panos Achlioptas}, \bibinfo{person}{Mira Shalah},
  \bibinfo{person}{Rapha{\"e}l Groscot}, {and} \bibinfo{person}{Leonidas~{J.}
  Guibas}.} \bibinfo{year}{2019}\natexlab{}.
\newblock \showarticletitle{Composite shape modeling via latent space
  factorization}. In \bibinfo{booktitle}{\emph{IEEE International Conference on
  Computer Vision (ICCV)}}. \bibinfo{pages}{8140--8149}.
\newblock


\bibitem[\protect\citeauthoryear{Dumoulin, Shlens, and Kudlur}{Dumoulin
  et~al\mbox{.}}{2017}]%
        {dumoulin2016learned}
\bibfield{author}{\bibinfo{person}{Vincent Dumoulin}, \bibinfo{person}{Jonathon
  Shlens}, {and} \bibinfo{person}{Manjunath Kudlur}.}
  \bibinfo{year}{2017}\natexlab{}.
\newblock \showarticletitle{A learned representation for artistic style}. In
  \bibinfo{booktitle}{\emph{International Conference on Learning
  Representations (ICLR)}}.
\newblock


\bibitem[\protect\citeauthoryear{Gal, Bermano, Zhang, and Cohen-Or}{Gal
  et~al\mbox{.}}{2020}]%
        {gal2020mrgan}
\bibfield{author}{\bibinfo{person}{Rinon Gal}, \bibinfo{person}{Amit Bermano},
  \bibinfo{person}{Hao Zhang}, {and} \bibinfo{person}{Daniel Cohen-Or}.}
  \bibinfo{year}{2020}\natexlab{}.
\newblock \showarticletitle{{MRGAN}: Multi-Rooted {3D} Shape Generation with
  Unsupervised Part Disentanglement}.
\newblock \bibinfo{journal}{\emph{arXiv preprint arXiv:2007.12944}}
  (\bibinfo{year}{2020}).
\newblock


\bibitem[\protect\citeauthoryear{Goodfellow, Pouget-Abadie, Mirza, Xu,
  Warde-Farley, Ozair, Courville, and Bengio}{Goodfellow et~al\mbox{.}}{2014}]%
        {goodfellow2014generative}
\bibfield{author}{\bibinfo{person}{Ian Goodfellow}, \bibinfo{person}{Jean
  Pouget-Abadie}, \bibinfo{person}{Mehdi Mirza}, \bibinfo{person}{Bing Xu},
  \bibinfo{person}{David Warde-Farley}, \bibinfo{person}{Sherjil Ozair},
  \bibinfo{person}{Aaron Courville}, {and} \bibinfo{person}{Yoshua Bengio}.}
  \bibinfo{year}{2014}\natexlab{}.
\newblock \showarticletitle{Generative adversarial nets}. In
  \bibinfo{booktitle}{\emph{Conference on Neural Information Processing Systems
  (NeurIPS)}}. \bibinfo{pages}{2672--2680}.
\newblock


\bibitem[\protect\citeauthoryear{Groueix, Fisher, Kim, Russell, and
  Aubry}{Groueix et~al\mbox{.}}{2018}]%
        {groueix2018papier}
\bibfield{author}{\bibinfo{person}{Thibault Groueix}, \bibinfo{person}{Matthew
  Fisher}, \bibinfo{person}{Vladimir~G. Kim}, \bibinfo{person}{Bryan~C.
  Russell}, {and} \bibinfo{person}{Mathieu Aubry}.}
  \bibinfo{year}{2018}\natexlab{}.
\newblock \showarticletitle{A papier-m{\^a}ch{\'e} approach to learning {3D}
  surface generation}. In \bibinfo{booktitle}{\emph{IEEE Conference on Computer
  Vision and Pattern Recognition (CVPR)}}. \bibinfo{pages}{216--224}.
\newblock


\bibitem[\protect\citeauthoryear{Guo, Xu, Yu, Liu, Dai, and Liu}{Guo
  et~al\mbox{.}}{2017}]%
        {guo2017real}
\bibfield{author}{\bibinfo{person}{Kaiwen Guo}, \bibinfo{person}{Feng Xu},
  \bibinfo{person}{Tao Yu}, \bibinfo{person}{Xiaoyang Liu},
  \bibinfo{person}{Qionghai Dai}, {and} \bibinfo{person}{Yebin Liu}.}
  \bibinfo{year}{2017}\natexlab{}.
\newblock \showarticletitle{Real-time geometry, albedo, and motion
  reconstruction using a single {RGB-D} camera}.
\newblock \bibinfo{journal}{\emph{ACM Transactions on Graphics (SIGGRAPH)}}
  \bibinfo{volume}{36}, \bibinfo{number}{3} (\bibinfo{year}{2017}),
  \bibinfo{pages}{32:1--32:13}.
\newblock


\bibitem[\protect\citeauthoryear{Hanocka, Metzer, Giryes, and Cohen-Or}{Hanocka
  et~al\mbox{.}}{2020}]%
        {hanocka2020point2mesh}
\bibfield{author}{\bibinfo{person}{Rana Hanocka}, \bibinfo{person}{Gal Metzer},
  \bibinfo{person}{Raja Giryes}, {and} \bibinfo{person}{Daniel Cohen-Or}.}
  \bibinfo{year}{2020}\natexlab{}.
\newblock \showarticletitle{{Point2Mesh}: A Self-Prior for Deformable Meshes}.
\newblock \bibinfo{journal}{\emph{ACM Transactions on Graphics (SIGGRAPH)}}
  \bibinfo{volume}{39}, \bibinfo{number}{4} (\bibinfo{year}{2020}),
  \bibinfo{pages}{126:1--126:12}.
\newblock


\bibitem[\protect\citeauthoryear{Hui, Xu, Xie, Qian, and Yang}{Hui
  et~al\mbox{.}}{2020}]%
        {hui2020progressive}
\bibfield{author}{\bibinfo{person}{Le Hui}, \bibinfo{person}{Rui Xu},
  \bibinfo{person}{Jin Xie}, \bibinfo{person}{Jianjun Qian}, {and}
  \bibinfo{person}{Jian Yang}.} \bibinfo{year}{2020}\natexlab{}.
\newblock \showarticletitle{Progressive point cloud deconvolution generation
  network}. In \bibinfo{booktitle}{\emph{European Conference on Computer Vision
  (ECCV)}}.
\newblock


\bibitem[\protect\citeauthoryear{Karras, Laine, and Aila}{Karras
  et~al\mbox{.}}{2019}]%
        {karras2019style}
\bibfield{author}{\bibinfo{person}{Tero Karras}, \bibinfo{person}{Samuli
  Laine}, {and} \bibinfo{person}{Timo Aila}.} \bibinfo{year}{2019}\natexlab{}.
\newblock \showarticletitle{A style-based generator architecture for generative
  adversarial networks}. In \bibinfo{booktitle}{\emph{IEEE Conference on
  Computer Vision and Pattern Recognition (CVPR)}}.
  \bibinfo{pages}{4401--4410}.
\newblock


\bibitem[\protect\citeauthoryear{Kim, Lee, Kang, Lee, and Kim}{Kim
  et~al\mbox{.}}{2020}]%
        {kim2020softflow}
\bibfield{author}{\bibinfo{person}{Hyeongju Kim}, \bibinfo{person}{Hyeonseung
  Lee}, \bibinfo{person}{Woo~Hyun Kang}, \bibinfo{person}{Joun~Yeop Lee}, {and}
  \bibinfo{person}{Nam~Soo Kim}.} \bibinfo{year}{2020}\natexlab{}.
\newblock \showarticletitle{{SoftFlow}: Probabilistic framework for normalizing
  flow on manifolds}. In \bibinfo{booktitle}{\emph{Conference on Neural
  Information Processing Systems (NeurIPS)}}.
\newblock


\bibitem[\protect\citeauthoryear{Klokov, Boyer, and Verbeek}{Klokov
  et~al\mbox{.}}{2020}]%
        {klokov2020discrete}
\bibfield{author}{\bibinfo{person}{Roman Klokov}, \bibinfo{person}{Edmond
  Boyer}, {and} \bibinfo{person}{Jakob Verbeek}.}
  \bibinfo{year}{2020}\natexlab{}.
\newblock \showarticletitle{Discrete point flow networks for efficient point
  cloud generation}. In \bibinfo{booktitle}{\emph{European Conference on
  Computer Vision (ECCV)}}.
\newblock


\bibitem[\protect\citeauthoryear{Knyaz, Kniaz, and Remondino}{Knyaz
  et~al\mbox{.}}{2018}]%
        {knyaz2018image}
\bibfield{author}{\bibinfo{person}{Vladimir~A. Knyaz},
  \bibinfo{person}{Vladimir~V Kniaz}, {and} \bibinfo{person}{Fabio Remondino}.}
  \bibinfo{year}{2018}\natexlab{}.
\newblock \showarticletitle{Image-to-voxel model translation with conditional
  adversarial networks}. In \bibinfo{booktitle}{\emph{European Conference on
  Computer Vision (ECCV)}}.
\newblock


\bibitem[\protect\citeauthoryear{Li, Dong, Peers, and Tong}{Li
  et~al\mbox{.}}{2017}]%
        {li2017modeling}
\bibfield{author}{\bibinfo{person}{Xiao Li}, \bibinfo{person}{Yue Dong},
  \bibinfo{person}{Pieter Peers}, {and} \bibinfo{person}{Xin Tong}.}
  \bibinfo{year}{2017}\natexlab{}.
\newblock \showarticletitle{Modeling surface appearance from a single
  photograph using self-augmented convolutional neural networks}.
\newblock \bibinfo{journal}{\emph{ACM Transactions on Graphics (SIGGRAPH)}}
  \bibinfo{volume}{36}, \bibinfo{number}{4} (\bibinfo{year}{2017}),
  \bibinfo{pages}{45:1--45:11}.
\newblock


\bibitem[\protect\citeauthoryear{Liu, Guo, Pan, Wang, Tong, and Liu}{Liu
  et~al\mbox{.}}{2021}]%
        {Liu2021MLS}
\bibfield{author}{\bibinfo{person}{Shi-Lin Liu}, \bibinfo{person}{Hao-Xiang
  Guo}, \bibinfo{person}{Hao Pan}, \bibinfo{person}{Pengshuai Wang},
  \bibinfo{person}{Xin Tong}, {and} \bibinfo{person}{Yang Liu}.}
  \bibinfo{year}{2021}\natexlab{}.
\newblock \showarticletitle{Deep Implicit Moving Least-Squares Functions for
  {3D} Reconstruction}. In \bibinfo{booktitle}{\emph{IEEE Conference on
  Computer Vision and Pattern Recognition (CVPR)}}.
\newblock


\bibitem[\protect\citeauthoryear{Loper, Mahmood, Romero, Pons-Moll, and
  Black}{Loper et~al\mbox{.}}{2015}]%
        {SMPL:2015}
\bibfield{author}{\bibinfo{person}{Matthew Loper}, \bibinfo{person}{Naureen
  Mahmood}, \bibinfo{person}{Javier Romero}, \bibinfo{person}{Gerard
  Pons-Moll}, {and} \bibinfo{person}{Michael~J. Black}.}
  \bibinfo{year}{2015}\natexlab{}.
\newblock \showarticletitle{{SMPL}: A skinned multi-person linear model}.
\newblock \bibinfo{journal}{\emph{ACM Transactions on Graphics (SIGGRAPH
  Asia)}} \bibinfo{volume}{34}, \bibinfo{number}{6} (\bibinfo{year}{2015}),
  \bibinfo{pages}{248:1--248:16}.
\newblock


\bibitem[\protect\citeauthoryear{Mao, Li, Xie, Lau, Wang, and Paul~Smolley}{Mao
  et~al\mbox{.}}{2017}]%
        {mao2017least}
\bibfield{author}{\bibinfo{person}{Xudong Mao}, \bibinfo{person}{Qing Li},
  \bibinfo{person}{Haoran Xie}, \bibinfo{person}{Raymond~Y.K. Lau},
  \bibinfo{person}{Zhen Wang}, {and} \bibinfo{person}{Stephen Paul~Smolley}.}
  \bibinfo{year}{2017}\natexlab{}.
\newblock \showarticletitle{Least squares generative adversarial networks}. In
  \bibinfo{booktitle}{\emph{IEEE International Conference on Computer Vision
  (ICCV)}}. \bibinfo{pages}{2794--2802}.
\newblock


\bibitem[\protect\citeauthoryear{Mescheder, Oechsle, Niemeyer, Nowozin, and
  Geiger}{Mescheder et~al\mbox{.}}{2019}]%
        {mescheder2019occupancy}
\bibfield{author}{\bibinfo{person}{Lars Mescheder}, \bibinfo{person}{Michael
  Oechsle}, \bibinfo{person}{Michael Niemeyer}, \bibinfo{person}{Sebastian
  Nowozin}, {and} \bibinfo{person}{Andreas Geiger}.}
  \bibinfo{year}{2019}\natexlab{}.
\newblock \showarticletitle{Occupancy networks: Learning {3D} reconstruction in
  function space}. In \bibinfo{booktitle}{\emph{IEEE Conference on Computer
  Vision and Pattern Recognition (CVPR)}}. \bibinfo{pages}{4460--4470}.
\newblock


\bibitem[\protect\citeauthoryear{Mo, Guerrero, Yi, Su, Wonka, Mitra, and
  Guibas}{Mo et~al\mbox{.}}{2019}]%
        {mo2019structurenet}
\bibfield{author}{\bibinfo{person}{Kaichun Mo}, \bibinfo{person}{Paul
  Guerrero}, \bibinfo{person}{Li Yi}, \bibinfo{person}{Hao Su},
  \bibinfo{person}{Peter Wonka}, \bibinfo{person}{Niloy Mitra}, {and}
  \bibinfo{person}{Leonidas~J. Guibas}.} \bibinfo{year}{2019}\natexlab{}.
\newblock \showarticletitle{Structure{N}et: Hierarchical graph networks for
  3{D} shape generation}.
\newblock \bibinfo{journal}{\emph{ACM Transactions on Graphics (SIGGRAPH
  Asia)}} \bibinfo{volume}{38}, \bibinfo{number}{6} (\bibinfo{year}{2019}),
  \bibinfo{pages}{242:1--242:19}.
\newblock


\bibitem[\protect\citeauthoryear{Mo, Wang, Yan, and Guibas}{Mo
  et~al\mbox{.}}{2020}]%
        {mo2020pt2pc}
\bibfield{author}{\bibinfo{person}{Kaichun Mo}, \bibinfo{person}{He Wang},
  \bibinfo{person}{Xinchen Yan}, {and} \bibinfo{person}{Leonidas~J. Guibas}.}
  \bibinfo{year}{2020}\natexlab{}.
\newblock \showarticletitle{{PT2PC}: Learning to generate 3{D} point cloud
  shapes from part tree conditions}. In \bibinfo{booktitle}{\emph{European
  Conference on Computer Vision (ECCV)}}.
\newblock


\bibitem[\protect\citeauthoryear{Park, Florence, Straub, Newcombe, and
  Lovegrove}{Park et~al\mbox{.}}{2019}]%
        {park2019deepsdf}
\bibfield{author}{\bibinfo{person}{Jeong~Joon Park}, \bibinfo{person}{Peter
  Florence}, \bibinfo{person}{Julian Straub}, \bibinfo{person}{Richard
  Newcombe}, {and} \bibinfo{person}{Steven Lovegrove}.}
  \bibinfo{year}{2019}\natexlab{}.
\newblock \showarticletitle{{DeepSDF}: Learning continuous signed distance
  functions for shape representation}. In \bibinfo{booktitle}{\emph{IEEE
  Conference on Computer Vision and Pattern Recognition (CVPR)}}.
  \bibinfo{pages}{165--174}.
\newblock


\bibitem[\protect\citeauthoryear{Qi, Su, Mo, and Guibas}{Qi
  et~al\mbox{.}}{2017a}]%
        {qi2016pointnet}
\bibfield{author}{\bibinfo{person}{Charles~R. Qi}, \bibinfo{person}{Hao Su},
  \bibinfo{person}{Kaichun Mo}, {and} \bibinfo{person}{Leonidas~J. Guibas}.}
  \bibinfo{year}{2017}\natexlab{a}.
\newblock \showarticletitle{{PointNet}: Deep learning on point sets for {3D}
  classification and segmentation}. In \bibinfo{booktitle}{\emph{IEEE
  Conference on Computer Vision and Pattern Recognition (CVPR)}}.
  \bibinfo{pages}{652--660}.
\newblock


\bibitem[\protect\citeauthoryear{Qi, Yi, Su, and Guibas}{Qi
  et~al\mbox{.}}{2017b}]%
        {qi2017pointnet++}
\bibfield{author}{\bibinfo{person}{Charles~R. Qi}, \bibinfo{person}{Li Yi},
  \bibinfo{person}{Hao Su}, {and} \bibinfo{person}{Leonidas~J. Guibas}.}
  \bibinfo{year}{2017}\natexlab{b}.
\newblock \showarticletitle{Point{N}et++: Deep hierarchical feature learning on
  point sets in a metric space}. In \bibinfo{booktitle}{\emph{Conference on
  Neural Information Processing Systems (NeurIPS)}}.
  \bibinfo{pages}{5099--5108}.
\newblock


\bibitem[\protect\citeauthoryear{Ramasinghe, Khan, Barnes, and
  Gould}{Ramasinghe et~al\mbox{.}}{2019}]%
        {ramasinghe2019spectral}
\bibfield{author}{\bibinfo{person}{Sameera Ramasinghe}, \bibinfo{person}{Salman
  Khan}, \bibinfo{person}{Nick Barnes}, {and} \bibinfo{person}{Stephen Gould}.}
  \bibinfo{year}{2019}\natexlab{}.
\newblock \showarticletitle{{Spectral-GANs} for high-resolution {3D}
  point-cloud generation}. In \bibinfo{booktitle}{\emph{IEEE/RSJ International
  Conference on Intelligent Robots and Systems (IROS)}}.
  \bibinfo{pages}{8169--8176}.
\newblock


\bibitem[\protect\citeauthoryear{Schonfeld, Schiele, and Khoreva}{Schonfeld
  et~al\mbox{.}}{2020}]%
        {schonfeld2020u}
\bibfield{author}{\bibinfo{person}{Edgar Schonfeld}, \bibinfo{person}{Bernt
  Schiele}, {and} \bibinfo{person}{Anna Khoreva}.}
  \bibinfo{year}{2020}\natexlab{}.
\newblock \showarticletitle{A {U-N}et based discriminator for generative
  adversarial networks}. In \bibinfo{booktitle}{\emph{IEEE Conference on
  Computer Vision and Pattern Recognition (CVPR)}}.
  \bibinfo{pages}{8207--8216}.
\newblock


\bibitem[\protect\citeauthoryear{Shu, Park, and Kwon}{Shu
  et~al\mbox{.}}{2019}]%
        {shu20193d}
\bibfield{author}{\bibinfo{person}{Dong~Wook Shu}, \bibinfo{person}{Sung~Woo
  Park}, {and} \bibinfo{person}{Junseok Kwon}.}
  \bibinfo{year}{2019}\natexlab{}.
\newblock \showarticletitle{{3D} point cloud generative adversarial network
  based on tree structured graph convolutions}. In
  \bibinfo{booktitle}{\emph{IEEE International Conference on Computer Vision
  (ICCV)}}. \bibinfo{pages}{3859--3868}.
\newblock


\bibitem[\protect\citeauthoryear{Sinha, Unmesh, Huang, and Ramani}{Sinha
  et~al\mbox{.}}{2017}]%
        {sinha2017surfnet}
\bibfield{author}{\bibinfo{person}{Ayan Sinha}, \bibinfo{person}{Asim Unmesh},
  \bibinfo{person}{Qixing Huang}, {and} \bibinfo{person}{Karthik Ramani}.}
  \bibinfo{year}{2017}\natexlab{}.
\newblock \showarticletitle{{SurfNet}: Generating {3D} shape surfaces using
  deep residual networks}. In \bibinfo{booktitle}{\emph{IEEE Conference on
  Computer Vision and Pattern Recognition (CVPR)}}.
  \bibinfo{pages}{6040--6049}.
\newblock


\bibitem[\protect\citeauthoryear{Smith and Meger}{Smith and Meger}{2017}]%
        {smith2017improved}
\bibfield{author}{\bibinfo{person}{Edward~J. Smith} {and}
  \bibinfo{person}{David Meger}.} \bibinfo{year}{2017}\natexlab{}.
\newblock \showarticletitle{Improved adversarial systems for {3D} object
  generation and reconstruction}. In \bibinfo{booktitle}{\emph{Conference on
  Robot Learning}}. PMLR, \bibinfo{pages}{87--96}.
\newblock


\bibitem[\protect\citeauthoryear{Sun, Wang, Liu, Siegel, and Sarma}{Sun
  et~al\mbox{.}}{2020}]%
        {sun2020pointgrow}
\bibfield{author}{\bibinfo{person}{Yongbin Sun}, \bibinfo{person}{Yue Wang},
  \bibinfo{person}{Ziwei Liu}, \bibinfo{person}{Joshua Siegel}, {and}
  \bibinfo{person}{Sanjay Sarma}.} \bibinfo{year}{2020}\natexlab{}.
\newblock \showarticletitle{{PointGrow}: Autoregressively learned point cloud
  generation with self-attention}. In \bibinfo{booktitle}{\emph{The IEEE Winter
  Conference on Applications of Computer Vision (WACV)}}.
  \bibinfo{pages}{61--70}.
\newblock


\bibitem[\protect\citeauthoryear{Waechter, Beljan, Fuhrmann, Moehrle, Kopf, and
  Goesele}{Waechter et~al\mbox{.}}{2017}]%
        {waechter2017virtual}
\bibfield{author}{\bibinfo{person}{Michael Waechter}, \bibinfo{person}{Mate
  Beljan}, \bibinfo{person}{Simon Fuhrmann}, \bibinfo{person}{Nils Moehrle},
  \bibinfo{person}{Johannes Kopf}, {and} \bibinfo{person}{Michael Goesele}.}
  \bibinfo{year}{2017}\natexlab{}.
\newblock \showarticletitle{Virtual rephotography: Novel view prediction error
  for {3D} reconstruction}.
\newblock \bibinfo{journal}{\emph{ACM Transactions on Graphics}}
  \bibinfo{volume}{36}, \bibinfo{number}{1} (\bibinfo{year}{2017}),
  \bibinfo{pages}{8:1--8:11}.
\newblock


\bibitem[\protect\citeauthoryear{Wang, Zhang, Li, Fu, Liu, and Jiang}{Wang
  et~al\mbox{.}}{2018}]%
        {wang2018pixel2mesh}
\bibfield{author}{\bibinfo{person}{Nanyang Wang}, \bibinfo{person}{Yinda
  Zhang}, \bibinfo{person}{Zhuwen Li}, \bibinfo{person}{Yanwei Fu},
  \bibinfo{person}{Wei Liu}, {and} \bibinfo{person}{Yu-Gang Jiang}.}
  \bibinfo{year}{2018}\natexlab{}.
\newblock \showarticletitle{{Pixel2Mesh}: Generating {3D} mesh models from
  single {RGB} images}. In \bibinfo{booktitle}{\emph{European Conference on
  Computer Vision (ECCV)}}. \bibinfo{pages}{52--67}.
\newblock


\bibitem[\protect\citeauthoryear{Wang, Sun, Liu, Sarma, Bronstein, and
  Solomon}{Wang et~al\mbox{.}}{2019}]%
        {wang2019dynamic}
\bibfield{author}{\bibinfo{person}{Yue Wang}, \bibinfo{person}{Yongbin Sun},
  \bibinfo{person}{Ziwei Liu}, \bibinfo{person}{Sanjay~E. Sarma},
  \bibinfo{person}{Michael~M. Bronstein}, {and} \bibinfo{person}{Justin~M.
  Solomon}.} \bibinfo{year}{2019}\natexlab{}.
\newblock \showarticletitle{Dynamic graph {CNN} for learning on point clouds}.
\newblock \bibinfo{journal}{\emph{ACM Transactions on Graphics}}
  \bibinfo{volume}{38}, \bibinfo{number}{5} (\bibinfo{year}{2019}),
  \bibinfo{pages}{146:1--146:12}.
\newblock


\bibitem[\protect\citeauthoryear{Wu, Wang, Xue, Sun, Freeman, and Tenenbaum}{Wu
  et~al\mbox{.}}{2017}]%
        {wu2017marrnet}
\bibfield{author}{\bibinfo{person}{Jiajun Wu}, \bibinfo{person}{Yifan Wang},
  \bibinfo{person}{Tianfan Xue}, \bibinfo{person}{Xingyuan Sun},
  \bibinfo{person}{Bill Freeman}, {and} \bibinfo{person}{Josh Tenenbaum}.}
  \bibinfo{year}{2017}\natexlab{}.
\newblock \showarticletitle{Marr{N}et: 3{D} shape reconstruction via 2.5{D}
  sketches}. In \bibinfo{booktitle}{\emph{Conference on Neural Information
  Processing Systems (NeurIPS)}}. \bibinfo{pages}{540--550}.
\newblock


\bibitem[\protect\citeauthoryear{Wu, Zhang, Xue, Freeman, and Tenenbaum}{Wu
  et~al\mbox{.}}{2016}]%
        {wu2016learning}
\bibfield{author}{\bibinfo{person}{Jiajun Wu}, \bibinfo{person}{Chengkai
  Zhang}, \bibinfo{person}{Tianfan Xue}, \bibinfo{person}{Bill Freeman}, {and}
  \bibinfo{person}{Josh Tenenbaum}.} \bibinfo{year}{2016}\natexlab{}.
\newblock \showarticletitle{Learning a probabilistic latent space of object
  shapes via 3{D} generative-adversarial modeling}. In
  \bibinfo{booktitle}{\emph{Conference on Neural Information Processing Systems
  (NeurIPS)}}. \bibinfo{pages}{82--90}.
\newblock


\bibitem[\protect\citeauthoryear{Wu, Zhuang, Xu, Zhang, and Chen}{Wu
  et~al\mbox{.}}{2020}]%
        {wu2020pq}
\bibfield{author}{\bibinfo{person}{Rundi Wu}, \bibinfo{person}{Yixin Zhuang},
  \bibinfo{person}{Kai Xu}, \bibinfo{person}{Hao Zhang}, {and}
  \bibinfo{person}{Baoquan Chen}.} \bibinfo{year}{2020}\natexlab{}.
\newblock \showarticletitle{{PQ-NET}: A generative part {Seq}2{S}eq network for
  3{D} shapes}. In \bibinfo{booktitle}{\emph{IEEE Conference on Computer Vision
  and Pattern Recognition (CVPR)}}. \bibinfo{pages}{829--838}.
\newblock


\bibitem[\protect\citeauthoryear{Wu, Song, Khosla, Yu, Zhang, Tang, and
  Xiao}{Wu et~al\mbox{.}}{2015}]%
        {wu20153d}
\bibfield{author}{\bibinfo{person}{Zhirong Wu}, \bibinfo{person}{Shuran Song},
  \bibinfo{person}{Aditya Khosla}, \bibinfo{person}{Fisher Yu},
  \bibinfo{person}{Linguang Zhang}, \bibinfo{person}{Xiaoou Tang}, {and}
  \bibinfo{person}{Jianxiong Xiao}.} \bibinfo{year}{2015}\natexlab{}.
\newblock \showarticletitle{3{D} {S}hape{N}ets: A deep representation for
  volumetric shapes}. In \bibinfo{booktitle}{\emph{IEEE Conference on Computer
  Vision and Pattern Recognition (CVPR)}}. \bibinfo{pages}{1912--1920}.
\newblock


\bibitem[\protect\citeauthoryear{Yang, Rosa, Markham, Trigoni, and Wen}{Yang
  et~al\mbox{.}}{2018}]%
        {yang2018dense}
\bibfield{author}{\bibinfo{person}{Bo Yang}, \bibinfo{person}{Stefano Rosa},
  \bibinfo{person}{Andrew Markham}, \bibinfo{person}{Niki Trigoni}, {and}
  \bibinfo{person}{Hongkai Wen}.} \bibinfo{year}{2018}\natexlab{}.
\newblock \showarticletitle{Dense {3D} object reconstruction from a single
  depth view}.
\newblock \bibinfo{journal}{\emph{IEEE Transactions Pattern Analysis \& Machine
  Intelligence}} \bibinfo{volume}{41}, \bibinfo{number}{12}
  (\bibinfo{year}{2018}), \bibinfo{pages}{2820--2834}.
\newblock


\bibitem[\protect\citeauthoryear{Yang, Huang, Hao, Liu, Belongie, and
  Hariharan}{Yang et~al\mbox{.}}{2019}]%
        {yang2019pointflow}
\bibfield{author}{\bibinfo{person}{Guandao Yang}, \bibinfo{person}{Xun Huang},
  \bibinfo{person}{Zekun Hao}, \bibinfo{person}{Ming-Yu Liu},
  \bibinfo{person}{Serge Belongie}, {and} \bibinfo{person}{Bharath Hariharan}.}
  \bibinfo{year}{2019}\natexlab{}.
\newblock \showarticletitle{{PointFlow}: 3{D} point cloud generation with
  continuous normalizing flows}. In \bibinfo{booktitle}{\emph{IEEE
  International Conference on Computer Vision (ICCV)}}.
  \bibinfo{pages}{4541--4550}.
\newblock


\bibitem[\protect\citeauthoryear{Yin, Chen, Huang, Cohen-Or, and Zhang}{Yin
  et~al\mbox{.}}{2019}]%
        {yin2019logan}
\bibfield{author}{\bibinfo{person}{Kangxue Yin}, \bibinfo{person}{Zhiqin Chen},
  \bibinfo{person}{Hui Huang}, \bibinfo{person}{Daniel Cohen-Or}, {and}
  \bibinfo{person}{Hao Zhang}.} \bibinfo{year}{2019}\natexlab{}.
\newblock \showarticletitle{{LOGAN}: Unpaired shape transform in latent
  overcomplete space}.
\newblock \bibinfo{journal}{\emph{ACM Transactions on Graphics (SIGGRAPH
  Asia)}} \bibinfo{volume}{38}, \bibinfo{number}{6} (\bibinfo{year}{2019}),
  \bibinfo{pages}{198:1--198:13}.
\newblock


\bibitem[\protect\citeauthoryear{Yin, Huang, Cohen-Or, and Zhang}{Yin
  et~al\mbox{.}}{2018}]%
        {yin2018p2p}
\bibfield{author}{\bibinfo{person}{Kangxue Yin}, \bibinfo{person}{Hui Huang},
  \bibinfo{person}{Daniel Cohen-Or}, {and} \bibinfo{person}{Hao Zhang}.}
  \bibinfo{year}{2018}\natexlab{}.
\newblock \showarticletitle{{P2P-Net}: Bidirectional point displacement net for
  shape transform}.
\newblock \bibinfo{journal}{\emph{ACM Transactions on Graphics (SIGGRAPH)}}
  \bibinfo{volume}{37}, \bibinfo{number}{4} (\bibinfo{year}{2018}),
  \bibinfo{pages}{152:1--152:13}.
\newblock


\bibitem[\protect\citeauthoryear{Yuan, Khot, Held, Mertz, and Hebert}{Yuan
  et~al\mbox{.}}{2018}]%
        {yuan2018pcn}
\bibfield{author}{\bibinfo{person}{Wentao Yuan}, \bibinfo{person}{Tejas Khot},
  \bibinfo{person}{David Held}, \bibinfo{person}{Christoph Mertz}, {and}
  \bibinfo{person}{Martial Hebert}.} \bibinfo{year}{2018}\natexlab{}.
\newblock \showarticletitle{{PCN}: Point completion network}. In
  \bibinfo{booktitle}{\emph{International Conference on 3D Vision (3DV)}}.
  \bibinfo{pages}{728--737}.
\newblock


\bibitem[\protect\citeauthoryear{Zheng, Yu, Wei, Dai, and Liu}{Zheng
  et~al\mbox{.}}{2019}]%
        {zheng2019deephuman}
\bibfield{author}{\bibinfo{person}{Zerong Zheng}, \bibinfo{person}{Tao Yu},
  \bibinfo{person}{Yixuan Wei}, \bibinfo{person}{Qionghai Dai}, {and}
  \bibinfo{person}{Yebin Liu}.} \bibinfo{year}{2019}\natexlab{}.
\newblock \showarticletitle{Deep{H}uman: 3{D} human reconstruction from a
  single image}. In \bibinfo{booktitle}{\emph{IEEE International Conference on
  Computer Vision (ICCV)}}. \bibinfo{pages}{7739--7749}.
\newblock


\bibitem[\protect\citeauthoryear{Zuffi, Kanazawa, Jacobs, and Black}{Zuffi
  et~al\mbox{.}}{2017}]%
        {Zuffi:CVPR:2017}
\bibfield{author}{\bibinfo{person}{Silvia Zuffi}, \bibinfo{person}{Angjoo
  Kanazawa}, \bibinfo{person}{David Jacobs}, {and} \bibinfo{person}{Michael~J.
  Black}.} \bibinfo{year}{2017}\natexlab{}.
\newblock \showarticletitle{{3D} {M}enagerie: modeling the {3D} shape and pose
  of animals}. In \bibinfo{booktitle}{\emph{IEEE Conference on Computer Vision
  and Pattern Recognition (CVPR)}}. \bibinfo{pages}{5524--5532}.
\newblock


\end{thebibliography}

\end{document}